\documentclass[letterpaper]{article} 
\usepackage{cuted}   
\usepackage{aaai25}  
\usepackage{times}  
\usepackage{helvet}  
\usepackage{courier}  
\usepackage[hyphens]{url}  
\usepackage{graphicx} 
\usepackage{natbib}  
\usepackage{caption} 
\usepackage{algorithm}
\usepackage{algorithmic}
\usepackage{latexsym}
\usepackage{booktabs}
\usepackage{multirow}
\usepackage{graphicx} 
\usepackage{amsmath} 
\usepackage{graphicx}
\usepackage{caption}
\usepackage{subcaption}
\usepackage{amssymb}

\usepackage[table]{xcolor} 
\usepackage{newfloat}
\usepackage{listings}
\usepackage{tabularx}
\DeclareCaptionStyle{ruled}{labelfont=normalfont,labelsep=colon,strut=off} 
\floatstyle{ruled}
\newfloat{listing}{tb}{lst}{}
\floatname{listing}{Listing}
\title{Not All Tokens and Heads Are Equally Important: Dual-Level Attention Intervention for Hallucination Mitigation}

\author{
    Lexiang Tang,
    Xianwei Zhuang,
    Bang Yang,
    Zhiyuan Hu,
    Hongxiang Li,
    Lu Ma,
    Jinghan Ru,
    Yuexian Zou\thanks{Corresponding author: Yuexian Zou.}
}
\affiliations{
    \textsuperscript{}Peking University, China\\
\{tanglexiang, xwzhuang, huzhiyuan, lihongxiang, maluqaq, rujinghan\}@stu.pku.edu.cn
    \{yangbang,zouyx\}@pku.edu.cn\\
}

\usepackage{bibentry}
\renewcommand{\footnote}[1]{}
\begin{document}

\maketitle

\begin{abstract}
Large vision-language models (LVLMs) have demonstrated impressive capabilities across diverse multimodal tasks, yet they remain highly susceptible to visual hallucinations (VH), often producing confident but inaccurate descriptions of visual content. Building on the insight that \textit{not all tokens and attention heads contribute equally to VH mitigation}, we introduce \textbf{VisFlow}, a lightweight and training-free framework that alleviates hallucinations by directly modulating attention patterns during inference. To address two primary challenges of VH, namely insufficient visual attention and the dominance of language priors, we identify three problematic attention behaviors in LVLMs: (1) disproportionate allocation of attention to uninformative or trailing visual tokens, (2) over-dependence on the previously generated token, and (3) excessive fixation on system prompts that hinders multimodal integration. To overcome these issues, VisFlow introduces a \textit{dual-level Attention Intervention}, consisting of \textit{Token-level Attention Intervention (TAI)}, which reinforces attention to salient visual regions, and \textit{Head-level Attention Intervention (HAI)}, which suppresses undue focus on system prompts and adjacent text tokens. Together, these interventions strengthen visual alignment while reducing linguistic bias. Extensive experiments across diverse models and benchmarks demonstrate that VisFlow effectively mitigates hallucinations with minimal computational overhead.

\end{abstract}

%

\section{Introduction}

\begin{figure*}[t]
  \centering
  \includegraphics[width=\textwidth, height=3.5cm]{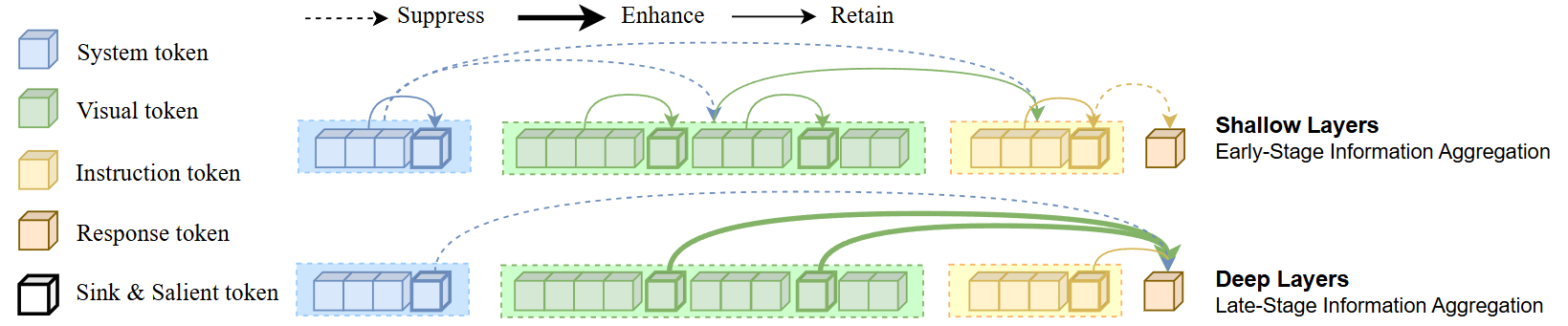} %
\caption{
Illustration of token-level attention distribution in VisFlow. We allocates stronger attention to visual salient token while avoiding over-attention to system prompts and prior text tokens, resulting in more balanced cross-modal alignment.
}
  \label{fig:visflow_info_flow}
\end{figure*}

Built upon the rapid progress of Large Language Models (LLMs)~\cite{yang2025qwen3,team2024qwen2,touvron2023llama,vicuna2023,ru2025we}, Large Vision-Language Models (LVLMs)~\cite{bai2025qwen2,wang2024qwen2,Chen2023ShikraUM,liu2023visual,ye2024mplug,bai2023qwen,chen2024internvl,li2023blip,chen2023minigpt,zhuang2025vargpt} have achieved strong performance across a wide range of multimodal understanding and generation tasks. An LVLM typically handles four types of tokens: (1) system prompts configuring model behavior, (2) visual tokens encoding image content, (3) instruction tokens representing user queries, and (4) response tokens as output text. Despite their capabilities, LVLMs often generate outputs misaligned with the visual input—a phenomenon known as visual hallucination (VH)~\cite{liu2024survey,leng2024mitigating,huang2024opera}, which poses risks in real-world and safety-critical applications.

\begin{figure*}[t]
    \centering
    \includegraphics[width=\textwidth, height=9cm]{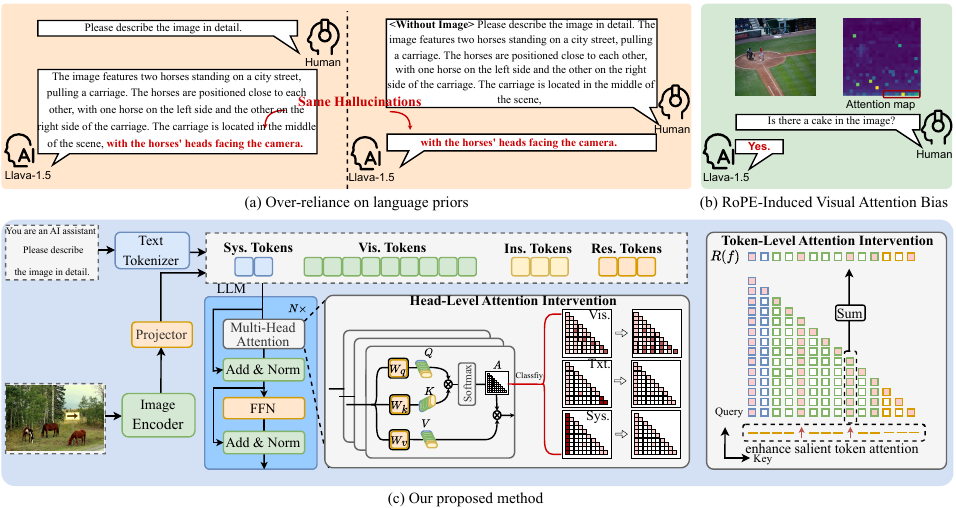} %
\caption{
Overview of visual hallucination causes and our solution.
(a) Linguistic Over-reliance: hallucinations caused by excessive dependence on language priors; 
(b) RoPE-induced Attention Bias: attention skewed toward image tokens near text tokens; 
(c) Our Method: mitigates these issues via Token-level Attention Intervention (TAI) to enhance focus on salient visual cues, and Head-level Attention Intervention (HAI) to suppress over-attention to system and nearby text tokens.
}
    \label{fig:method_and_bias}
\end{figure*}

Existing VH mitigation strategies fall into three categories: (1) \textit{Instruction tuning} with hallucination-aware datasets~\cite{jiang2024hallucination,sarkar2024data,chen2025perturbollava,yang2025mitigating}, which improves grounding but requires costly retraining; (2) \textit{Auxiliary modules} such as reranking or hallucination detection~\cite{manakul2023selfcheckgpt,yin2024woodpecker,chen2024halc}, which introduce additional latency and complexity; and (3) \textit{Decoding-time interventions}~\cite{huang2024opera,wang2024mitigating,fan2025improving,liu2024reducing,zhuang2025vasparse}, which are more efficient but often depend on contrastive decoding or external grounding tools, limiting scalability. In this work, we present VisFlow, a training-free framework that mitigates VH at inference by directly modulating attention patterns within the LVLM decoder. Building on prior analyses~\cite{liu2024survey,yin2025clearsight}, we focus on two primary contributing factors: (1) insufficient attention to visual information, and (2) over-reliance on language priors. These patterns are shown in Figures~\ref{fig:method_and_bias}(a)(b), and VisFlow's effectiveness in addressing them is illustrated in Figure~\ref{fig:case}.

Addressing the first issue, PAI~\cite{liu2024paying} enhances attention weights equally across all visual tokens, while VAR~\cite{kang2025see} introduces the concept of the visual sink token and reallocates redundant attention from the BOS and visual sink tokens to other visual tokens to improve visual alignment. However, these approaches lack precise targeting and may inadvertently reinforce cross-modal attention errors introduced by Rotary Positional Embedding (RoPE)~\cite{su2024roformer}. 

\textit{To overcome these limitations, we identified a critical insight: not all visual tokens are equally important for mitigating hallucination.} Leveraging this, we propose Token-Level Attention Intervention (TAI), which selectively enhances attention to these crucial visual tokens while also correcting for RoPE-induced bias. A detailed comparison of TAI's corrective effect is provided in the appendix C.1. To our knowledge, we are the first to identify critical visual tokens such as the visual sink and salient tokens based on intra-modal attention. Inspired by prior work~\cite{darcet2023vision,wang2023label}, which suggests that the visual sink token aggregates global semantic information from preceding tokens, we hypothesize that suppressing textual attention to the sink token may impair the model’s perception of global visual semantics. We validate this hypothesis through targeted intervention experiments on the POPE~\cite{li2023evaluating} benchmark, specifically focusing on the sink token. Unlike VAR, which redistributes attention away from it, TAI amplifies attention to the sink token on RoPE. Results in Figure~\ref{fig:ablation_enhance_token} support the visual sink token's crucial role.

Addressing the second issue, contrastive decoding paradigms such as VCD~\cite{leng2024mitigating} and ICD~\cite{wang2024mitigating}  mitigate hallucination by introducing perturbations to the visual or textual inputs, thereby increasing the uncertainty of model outputs and producing contrastive distributions dominated by linguistic priors. These contrastive distributions are then subtracted from the original ones to suppress hallucinations. However, generating these contrastive distributions requires multiple forward passes, leading to significant inference overhead. 

\textit{To overcome this, we introduce Head-Level Attention Intervention (HAI), a method built on a critical insight: not all heads are equally important for mitigating hallucination.} Our approach is the first to explicitly identify text attention heads correlated with linguistic priors and reduce their abnormal attention on text tokens. Since HAI directly modifies the original attention distribution, it provides a clear efficiency advantage over contrastive decoding methods, which require multi-pass decoding~\cite{leng2024mitigating,park2025convis,huo2024self,wang2024mitigating,an2025mitigating}. Our method avoids their computational burden, Tokens Per Second (TPS) comparison results are shown in Fig.~\ref{fig:tps}.

Our main contributions are as follows. (1) We analyze VH in LVLMs through the lens of attention and information flow, identifying different types of visual tokens and attention heads associated with hallucination. (2) We introduce VisFlow, a training-free and efficient inference-time framework incorporating TAI and HAI to enhance visual alignment and suppress hallucinations. (3) Comprehensive experiments demonstrate that VisFlow outperforms existing methods in both effectiveness and efficiency.

\section{RELATED WORK}

\subsubsection{Large Vision-Language Model (LVLM)}
LVLMs combine large language models (LLMs) with visual encoders to handle multimodal inputs. Early approaches, such as the LLaVA series~\cite{liu2024improved,liu2023visual}, align visual features with LLMs via linear projections and enhance performance through instruction tuning. Other works like BLIP-2~\cite{li2023blip}, MiniGPT-4~\cite{chen2023minigpt}, and InstructBLIP~\cite{liu2023visual} introduce query transformers (e.g., Q-former~\cite{li2023blip}) to extract instruction-aware visual features for improved efficiency. Recent models such as Qwen2.5-VL~\cite{bai2025qwen2,bai2023qwen}, mPLUG-Owl2~\cite{ye2024mplug}, and InternVL2.5~\cite{chen2024expanding,chen2024internvl} further optimize architectures, training, and data pipelines.

\subsubsection{Visual Hallucination Mitigation}

Efforts to mitigate visual hallucination (VH) in LVLMs fall into three categories: (1) \textit{Instruction tuning}~\cite{gunjal2024detecting, jiang2024hallucination}, which improves grounding but requires task-specific data and costly retraining; (2) \textit{Auxiliary analysis}~\cite{manakul2023selfcheckgpt, chen2024halc, yin2024woodpecker, an2024agla}, which adds inference-time modules at the cost of latency; and (3) \textit{Decoding-time interventions}~\cite{huang2024opera, wang2024mitigating, fan2025improving, liu2024reducing, zhuang2025vasparse}, which intervene during generation and are more efficient than retraining or auxiliary modules, but often rely on contrastive decoding or external grounding tools, increasing inference latency. Recent attention-based interventions~\cite{yang2025understanding, kang2025see, yin2025clearsight} directly modify attention distributions during inference, avoiding extra decoding steps, but often lack fine-grained targeting, focusing either on attention heads~\cite{yang2025understanding, yin2025clearsight} or token importance~\cite{zou2024look, kang2025see}. Our method \textbf{VisFlow} integrates Token Attention Intervention (TAI) and Head Attention Intervention (HAI) to jointly identify salient tokens and critical heads, enabling precise attention correction with minimal computational overhead.


\subsubsection{Information Flow in MLLMs}
Zhang et al.~\cite{zhang2024cross} provide an empirical visualization of cross-modal information flow in the decoder of LVLMs, outlining a three-stage process from global visual feature integration to final output generation. However, their analysis overlooks interactions among visual tokens. We address this gap by analyzing attention between visual tokens. Additionally, CCA-LLaVA~\cite{xing2024mitigating} attributes VH to disrupted token-level information flow, particularly the long-range decay introduced by RoPE. Building on this insight, as shown in Figure~\ref{fig:visflow_info_flow} and~\ref{fig:attention_combined}, we analyze the token-level information flow and attention distribution in LVLMs. Our proposed TAI module mitigates RoPE-induced VH.

\section{METHOD}
This section analyzes token-level information flow and attention distribution in MLLMs, and introduces our dual-level intervention methods.

\subsubsection{Measuring Information Flow in Multimodal Tokens}  
To analyze how MLLMs utilize visual information and why they may overly rely on language priors, we adopt the saliency technique~\cite{simonyan2013deep}, to highlight critical token interactions within the attention mechanism. The saliency score is computed by taking the Hadamard product of the attention scores A and their gradients as follows:
\begin{equation}
    I_{l} = \left| \sum_h A_{h,l} \odot \frac{\partial \mathcal{L}(x)}{\partial A_{h,l}} \right|,
\end{equation}
where $A_{h,l}$ is the attention matrix from the $h$-th head in the $l$-th layer, and $\mathcal{L}(x)$ denotes the task loss. The saliency matrix $I_l$ aggregates all heads to reflect the contribution of token $j$ to token $i$ in layer $l$.

\begin{figure}[t]
    \centering
    \begin{subfigure}[t]{0.48\linewidth}
        \centering
        \includegraphics[width=\linewidth]{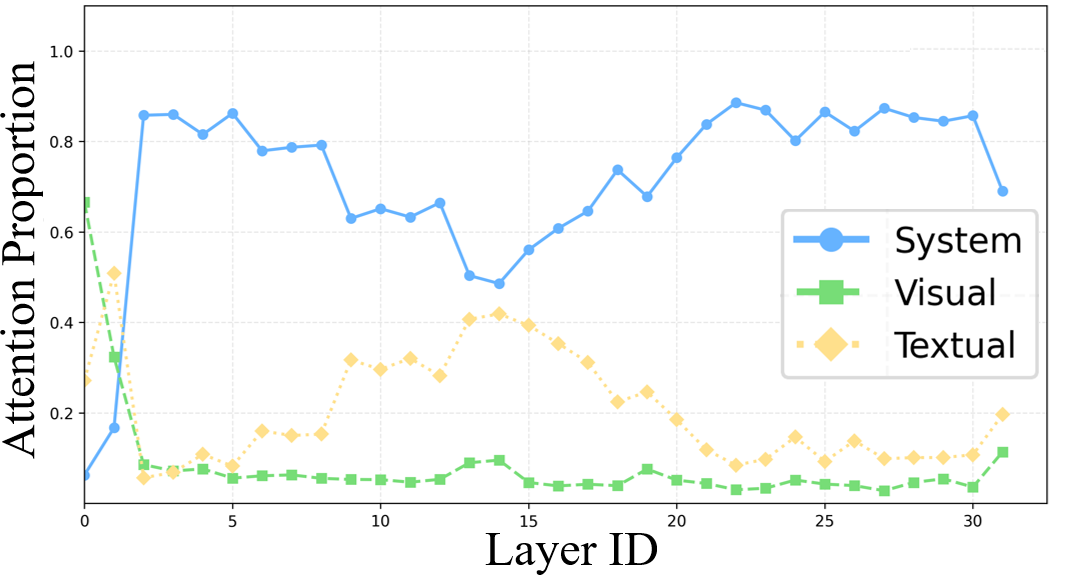}
        \caption{Attention distribution.}
        \label{fig:attention_distribution}
    \end{subfigure}
    \hfill
    \begin{subfigure}[t]{0.48\linewidth}
        \centering
        \includegraphics[width=\linewidth]{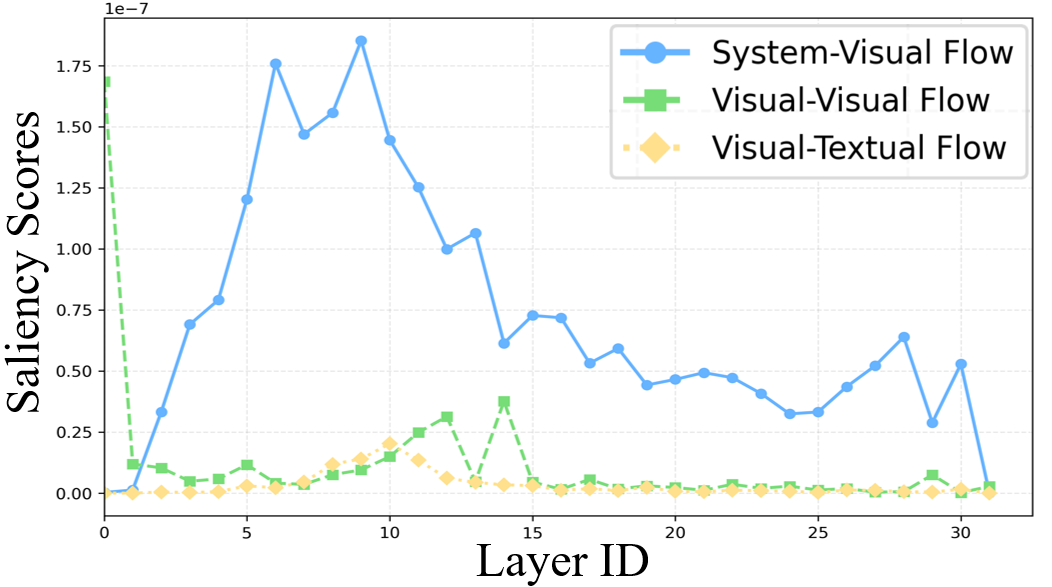}
        \caption{Salience scores.}
        \label{fig:attention_flow}
    \end{subfigure}
    \caption{Layer-wise token attention and information flow analysis on 500 MSCOCO samples from the POPE.}
    \label{fig:attention_combined}
\end{figure}

To quantify directional information flow between token groups (e.g., system, visual, textual), we define:
\begin{align}
    S_{ab} &= \frac{\sum_{(i, j) \in C_{ab}} I(i, j)}{|C_{ab}|}, \\
    C_{ab} &= \{(i, j) : i \in \mathcal{A},\ j \in \mathcal{B}\},
\end{align}
where $I(i, j)$ measures information flow from token $j$ to token $i$, and $C_{ab}$ is the set of all such directed token pairs. For example, $S_{sv}$ measures system-to-visual flow, $S_{vv}$ captures intra-visual flow (optionally constrained by $i \geq j$), and $S_{vt}$ quantifies visual-to-text transfer.

Figures~\ref{fig:attention_distribution} and~\ref{fig:attention_flow} consistently show biased attention patterns: insufficient attention and information flow on visual tokens and abnormally high focus on system prompts, suggesting impaired visual alignment.

\subsection{Token-Level Attention Intervention}

\begin{figure*}[htbp]
  \centering
  \begin{subfigure}{0.195\linewidth}
    \includegraphics[width=\linewidth]{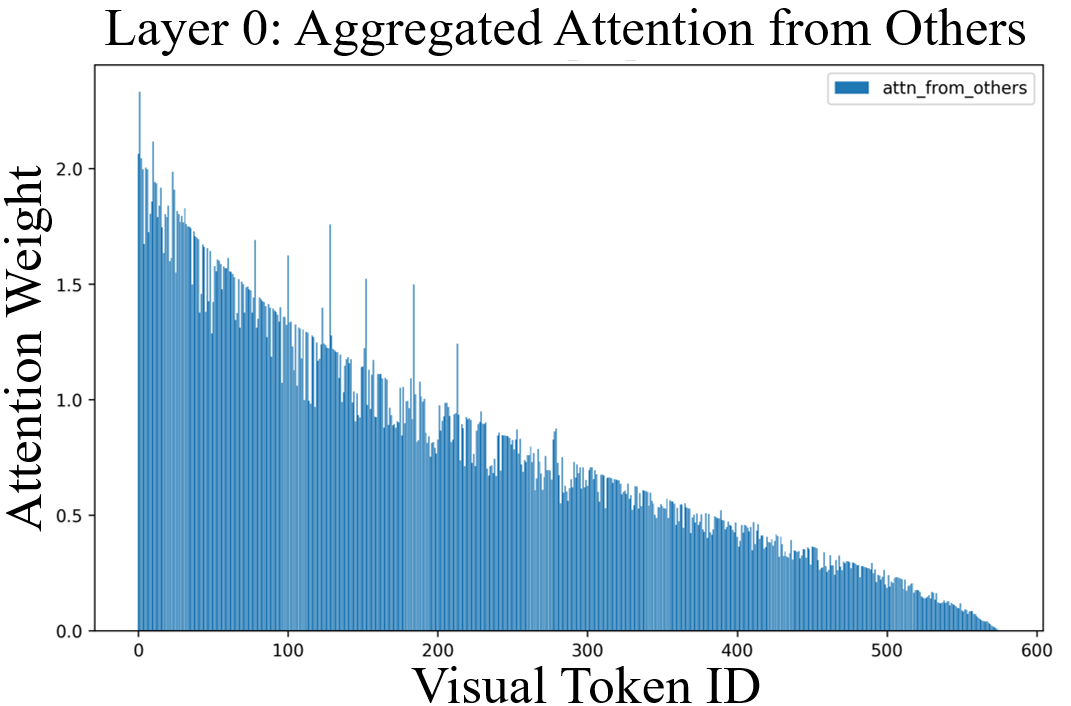}
    \caption{Layer 0.}
    \label{fig:short-a}
  \end{subfigure}
  \hfill
  \begin{subfigure}{0.195\linewidth}
    \includegraphics[width=\linewidth]{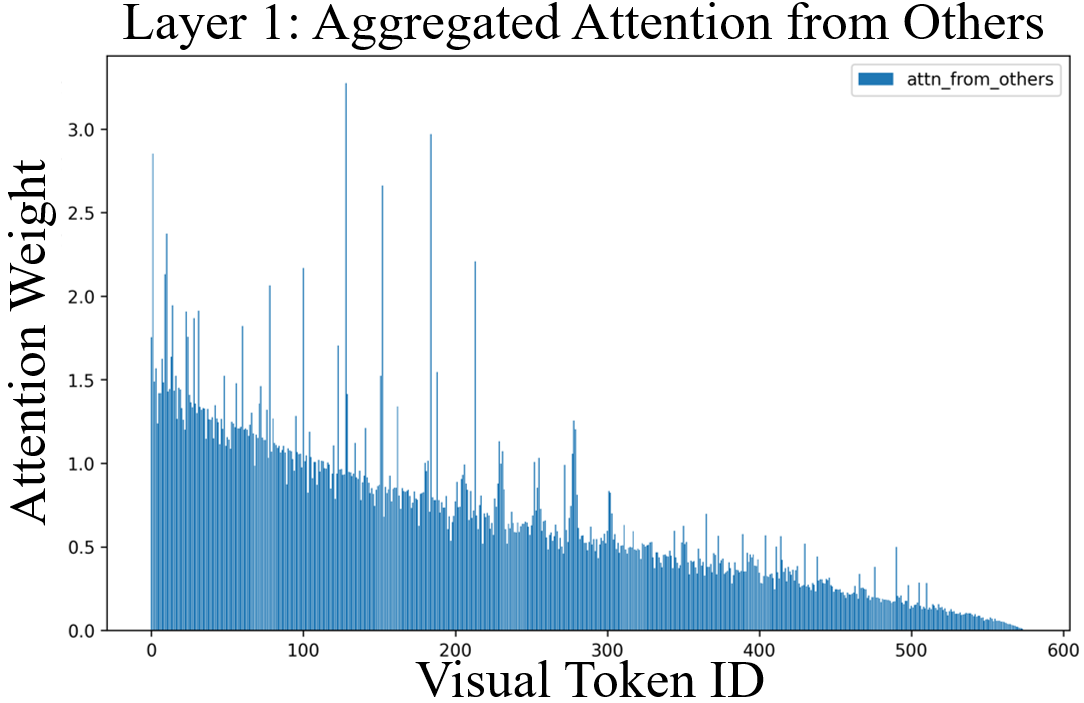}
    \caption{Layer 1.}
    \label{fig:short-b}
  \end{subfigure}
  \hfill
  \begin{subfigure}{0.195\linewidth}
    \includegraphics[width=\linewidth]{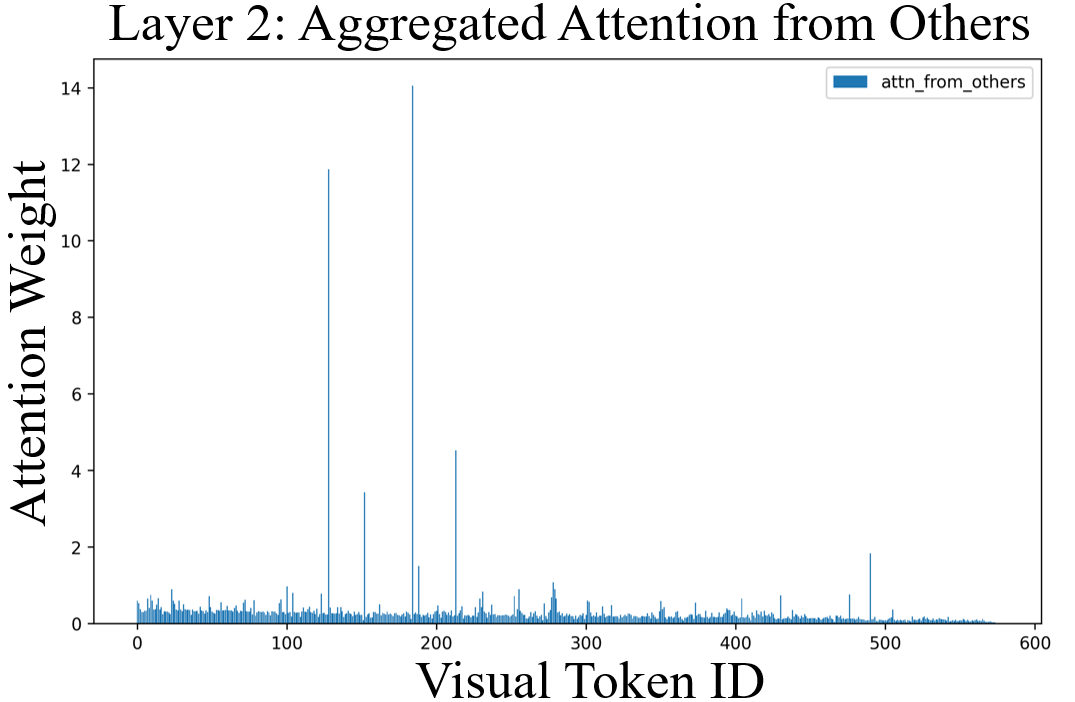}
    \caption{Layer 2.}
    \label{fig:short-c}
  \end{subfigure}
  \begin{subfigure}{0.195\linewidth}
    \includegraphics[width=\linewidth]{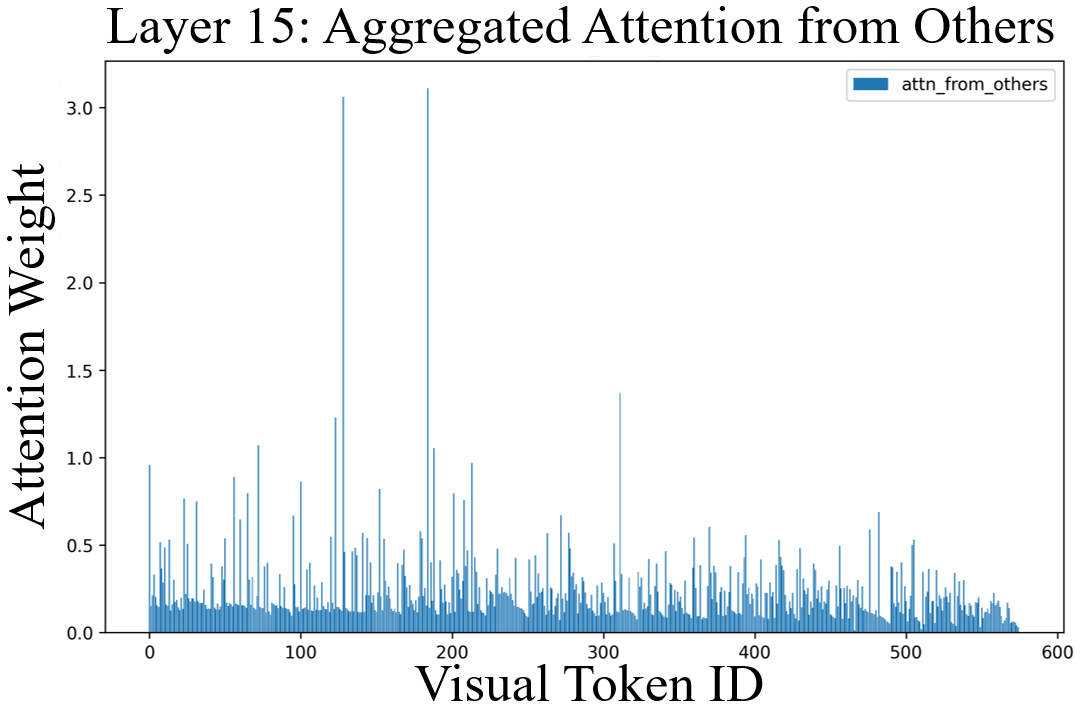}
    \caption{Layer 15.}
    \label{fig:short-c}
  \end{subfigure}
    \begin{subfigure}{0.195\linewidth}
    \includegraphics[width=\linewidth]{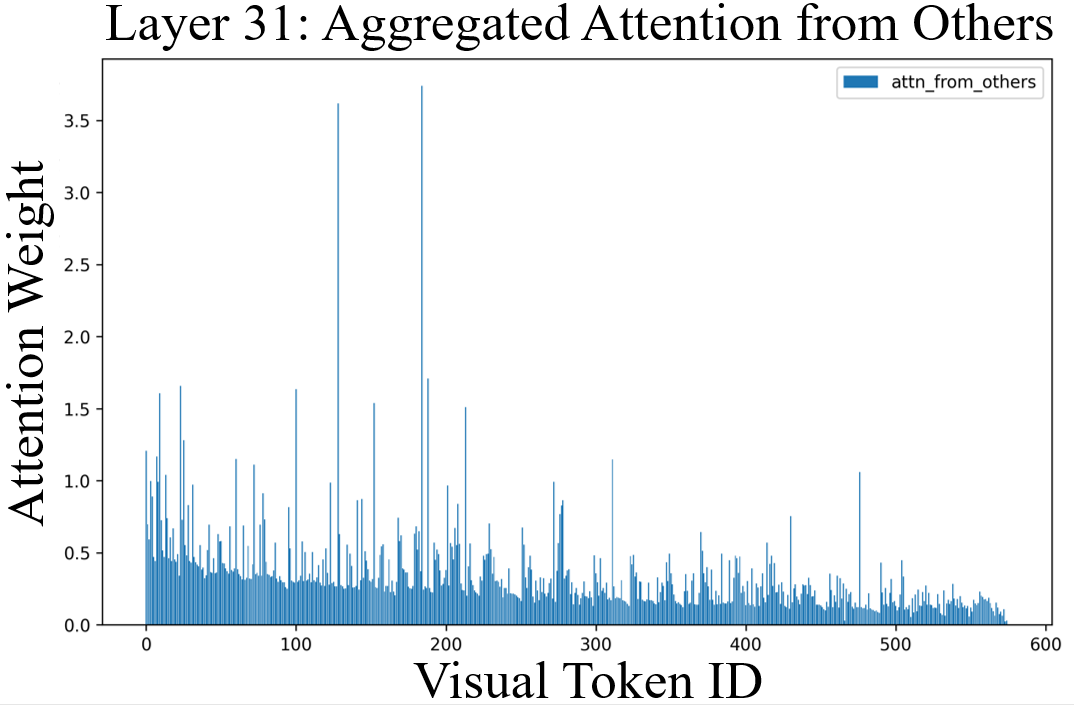}
    \caption{Layer 31.}
    \label{fig:short-c}
  \end{subfigure}
  \caption{Visualization of intra-visual reception score $R(j)$ among visual tokens across decoder Layers.}
  \label{fig:short}
\end{figure*}
\subsubsection{Visual Sink and Salient Token Identification}
To address misaligned visual attention, partly due to RoPE-induced position bias, we identify visual-salient tokens by analyzing intra-visual attention patterns. We define the \textit{reception score} $R(j)$ of a visual token $j \in \mathcal{I}_{\text{vis}}$ as the total attention it receives from other visual tokens across all heads in a decoder layer:

\begin{equation}
    \text{R}(j) = \frac{1}{H} \sum_{h=1}^{H} \sum_{i \in \mathcal{I}_{\text{vis}} \setminus \{j\}} \mathbf{A}_\ell^{(h)}[i, j],
\end{equation}

where $\mathbf{A}_\ell^{(h)}$ is the attention matrix of head $h$ in layer $\ell$, and $H$ is the number of heads. Figure~\ref{fig:short} shows the distribution of $R(j)$ across layers, revealing uneven attention among visual tokens. As shown in Figure~\ref{fig:short-a} and Figure~\ref{fig:short-b}, early layers exhibit globally distributed interactions among visual tokens, with clear bias introduced by causal attention. However, starting from layer index 2, distinct visual tokens with significantly higher $R(j)$ begin to emerge. These tokens correspond to the visually salient or sink tokens we aim to identify. To extract them, we select tokens whose $R(j)$ exceed a fraction $\tau$ of the maximum score:

\begin{multline}
    \mathcal{I}_{\text{thres}}(\tau) =
    \left\{ j \in \mathcal{I}_{\text{vis}} \;\middle|\;
    \text{R}(j) > \tau \cdot \max_{k \in \mathcal{I}_{\text{vis}}} \text{R}(k) \right\},
\end{multline}

where a small $\tau$ (e.g., $\tfrac{1}{20}$) selects \textit{salient tokens}, while a larger $\tau$ (e.g., $\tfrac{1}{2}$) indicates \textit{sink tokens}.




\begin{figure}
    \centering
    \begin{subfigure}[t]{0.32\linewidth}
        \centering
        \includegraphics[width=\linewidth]{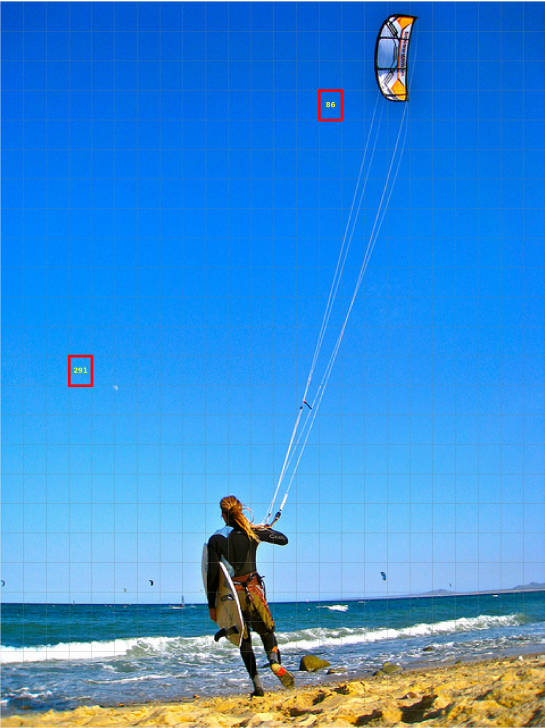}
        \caption{Sink token}
        \label{fig:sink_token}
    \end{subfigure}%
    \hfill
    \begin{subfigure}[t]{0.32\linewidth}
        \centering
        \includegraphics[width=\linewidth]{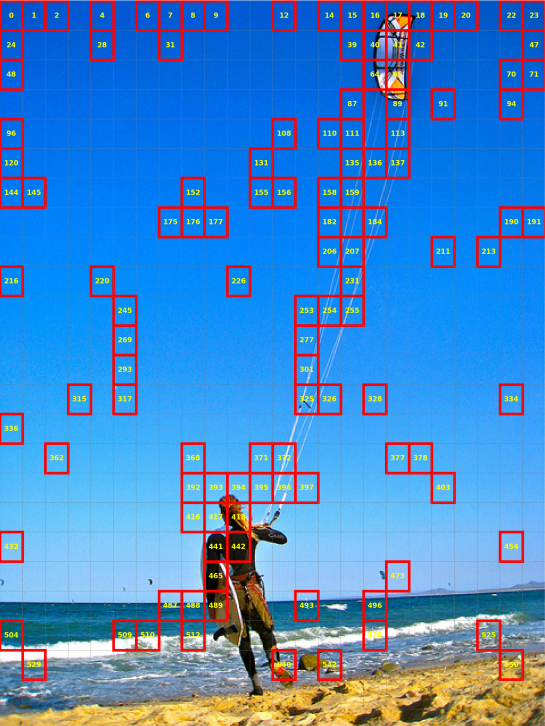}
        \caption{Salient token}
        \label{fig:salient_token}
    \end{subfigure}
    \begin{subfigure}[t]{0.323\linewidth}
        \centering
        \includegraphics[width=\linewidth]{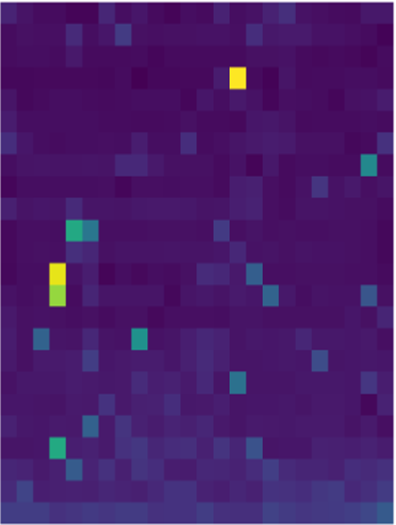}
        \caption{Attention map}
        \label{fig:attention_map}
    \end{subfigure}%
    \hfill
    \caption{Visualization of visual attention in LLaVA-1.5-7B. 
(a) Visual Sink Token: tokens that absorb much attention but lacks semantic contribution; 
(b) Visual Salient Token: tokens that align with meaningful visual regions critical for grounding; 
(c) Attention distribution over visual tokens. }
    \label{fig:three_subfigs}
\end{figure}
As illustrated in Figure~\ref{fig:three_subfigs}, we visualize the identified sink and salient tokens for a given image, alongside the full attention map, to highlight the discrepancy between attention magnitude and true visual relevance. More visualization cases are provided in the Appendix C.1.





\subsubsection{Enhancing Intra-Visual Salient Token Attention}
To improve visual alignment and suppress attention bias introduced by RoPE encoding, we modify the attention weights from instruction tokens to visual tokens at each decoder layer $\ell$ and head $h$. Specifically, we amplify attention toward salient regions and attenuate it for semantically meaningless sink tokens. The adjusted attention weights are defined as:

\begin{equation}
    A^{\ell,h}_{i,j} =
    \begin{cases}
        k \cdot A^{\ell,h}_{i,j}, & \text{if } i \in \mathcal{I}_{\text{txt}},\ j \in \mathcal{I}_{\text{salient}}^\ell \\
        \delta \cdot A^{\ell,h}_{i,j}, & \text{if } i \in \mathcal{I}_{\text{txt}},\ j \in \mathcal{I}_{\text{sink}}^\ell
    \end{cases}
\end{equation}

where $k > 1$ is a scaling factor that enhances attention to salient visual tokens, and $\delta < 1$ is a decay factor that suppresses attention to sink tokens.


To ensure the attention weights sum to 1, the modified weights are re-normalized as follows:

\begin{equation}
    A^{\ell,h}_{i,j} = \frac{A^{\ell,h}_{i,j}}{\sum\limits_{j} A^{\ell,h}_{i,j}}, \quad \text{if } i \in \mathcal{I}_{\text{txt}}.
\end{equation}

This strategy encourages the model to attend more effectively to informative visual evidence while reducing focus on irrelevant regions.




\subsection{Head-Level Attention Intervention}
\begin{figure*}[t]
    \centering
    \begin{subfigure}{0.32\linewidth}
        \centering
        \includegraphics[width=\linewidth]{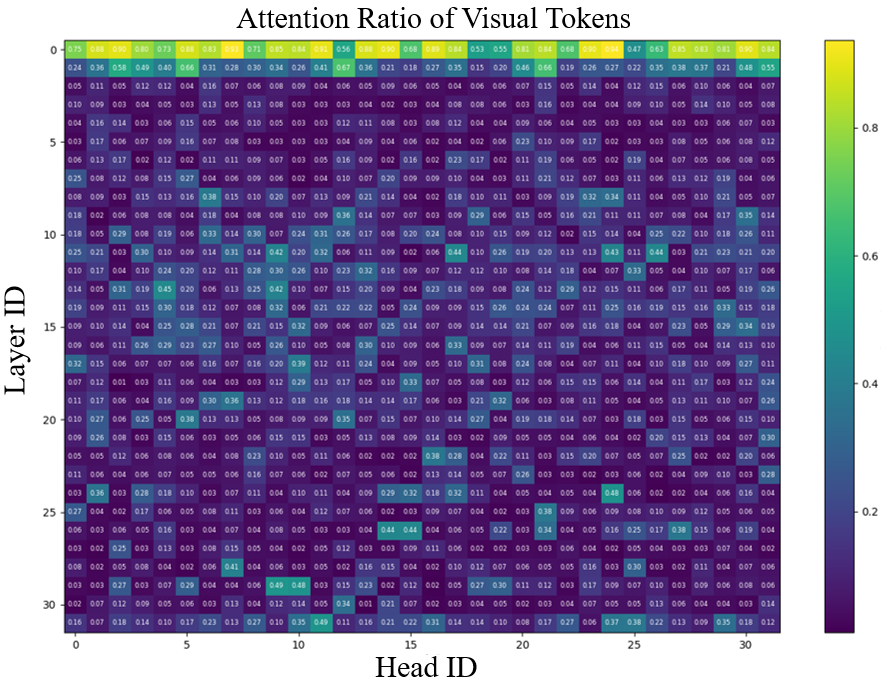}
        \caption{Visual attention}
        \label{fig:enc_vis_attn_heads_distribution}
    \end{subfigure}
    \begin{subfigure}{0.32\linewidth}
        \centering
        \includegraphics[width=\linewidth]{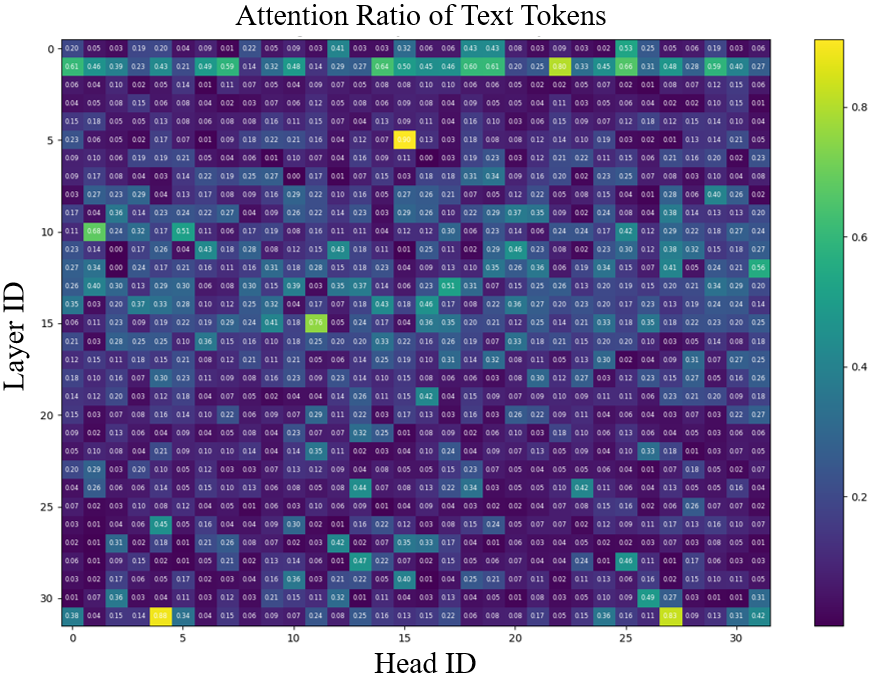}
        \caption{Text attention}
        \label{fig:enc_txt_attn_heads_distribution}
    \end{subfigure}
    \begin{subfigure}{0.32\linewidth}
        \centering
        \includegraphics[width=\linewidth]{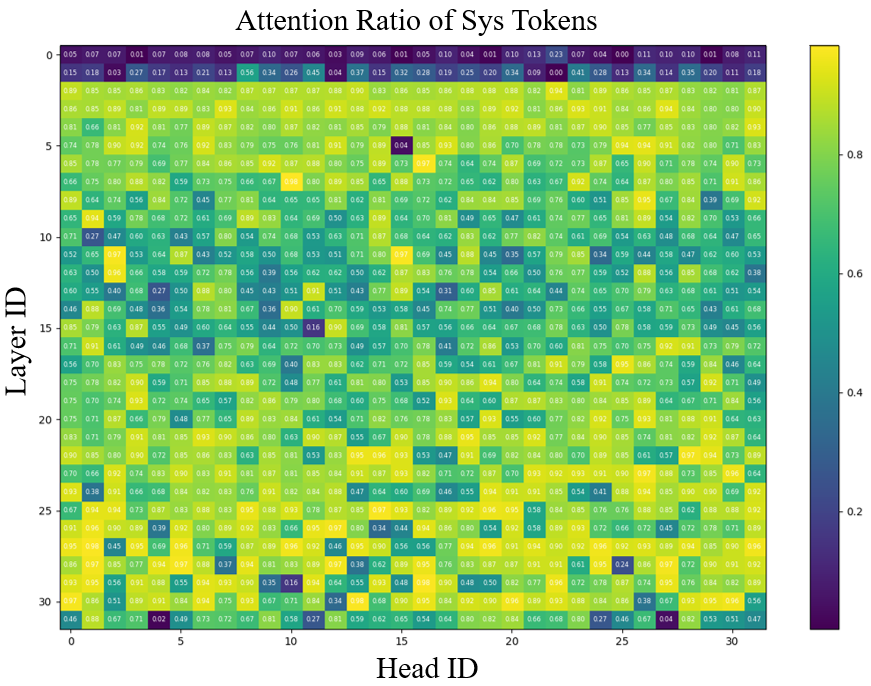}
        \caption{System attention}
        \label{fig:enc_sys_attn_heads_distribution}
    \end{subfigure}
    \caption{Layer-wise heatmaps of attention weights across all heads, showing preference for visual, text, and system tokens.}
    \label{fig:all_attn_heads_distribution}
\end{figure*}

\begin{figure}[t]
    \centering
    \begin{subfigure}[t]{0.32\linewidth}
        \centering
        \includegraphics[width=\linewidth]{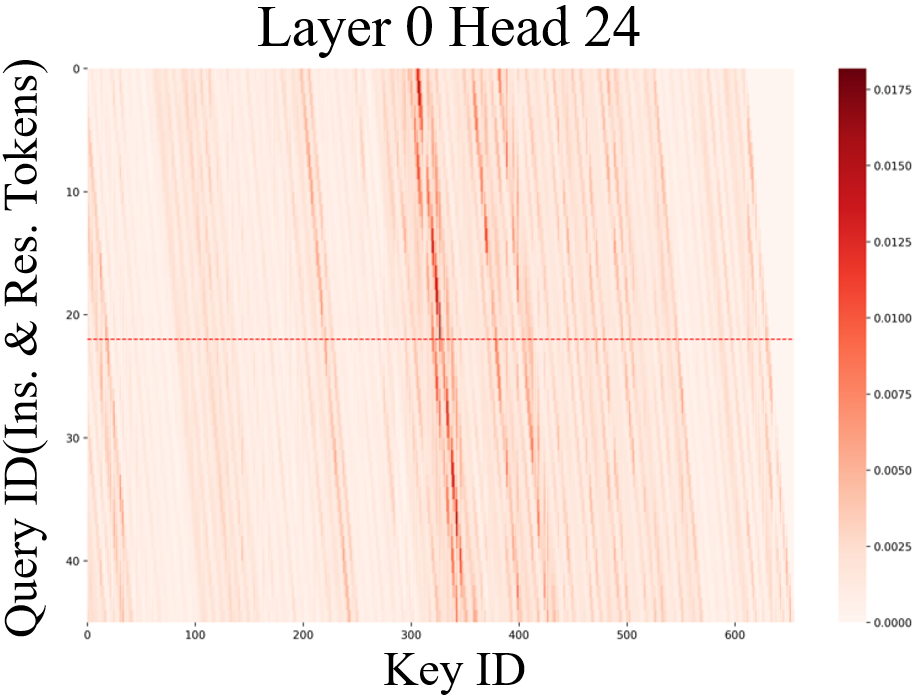}
        \caption{Visual head}
        \label{fig:visual_attn_head}
    \end{subfigure}
    \begin{subfigure}[t]{0.32\linewidth}
        \centering
        \includegraphics[width=\linewidth]{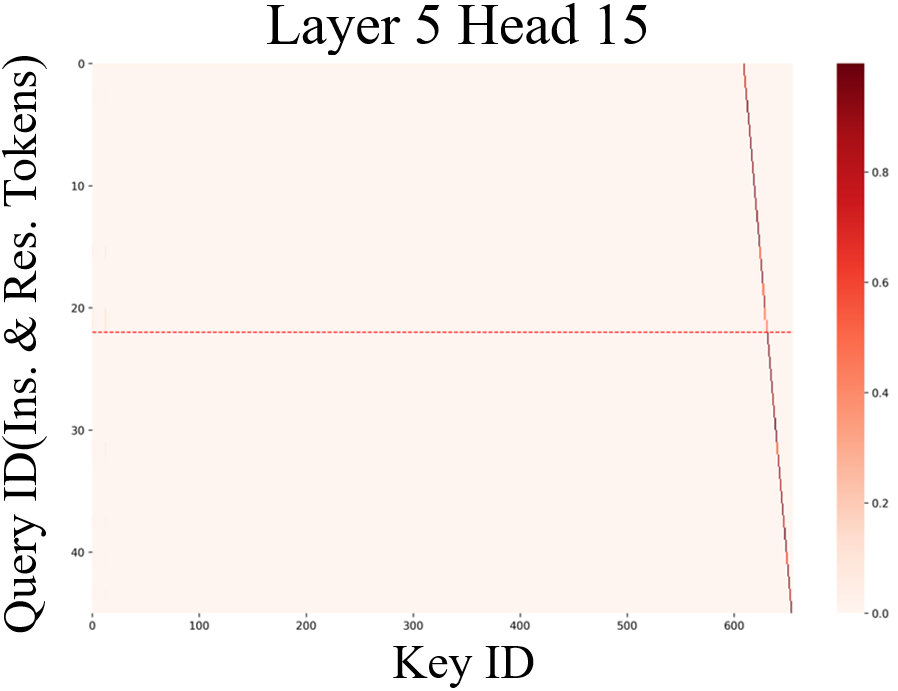}
        \caption{Text head}
        \label{fig:text_attn_head}
    \end{subfigure}
    \begin{subfigure}[t]{0.32\linewidth}
        \centering
        \includegraphics[width=\linewidth]{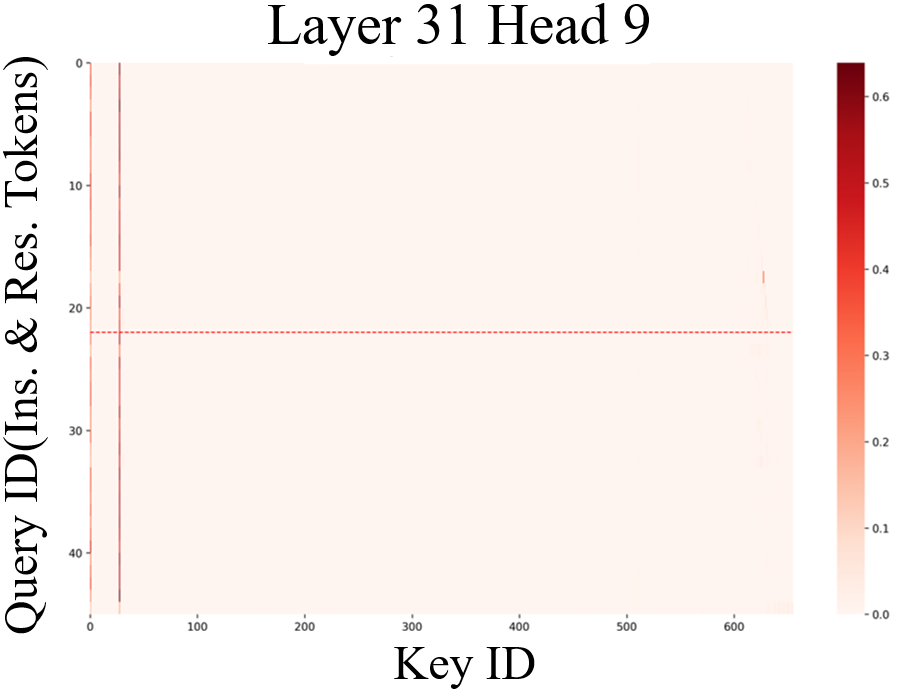}
        \caption{System head}
        \label{fig:system_attn_head}
    \end{subfigure}
    \caption{
    Attention maps from representative heads across modalities. Attention is visualized from instruction tokens (prefill stage) and response tokens (decode stage) to all key tokens. Red line separates the two phases.
    }
    \label{fig:attn_patterns_heads}
\end{figure}
\subsubsection{Attention Head Type Identification}
Prior studies~\cite{sun2024redeep,tang2025intervening,
michel2019sixteen,yang2025understanding} have shown that attention heads in large models often assume specialized functions. To investigate how different token types shape attention in decoder, we categorize heads into three types:
(1) \textit{Visual-sensitive}, focusing on visual tokens;
(2) \textit{System-dominant}, attending mainly to system prompts;
(3) \textit{Text-dominant}, aligned with instruction and response tokens.

\textbf{Identifying Visual-Sensitive Heads.} For each head $h$ in decoder layer $\ell$, we measure its attention to visual tokens as:



\begin{equation}
    A_{\text{vis}}^{\ell,h} = \sum_{i \in \mathcal{I}_{\text{txt}}} \sum_{j \in 
    \mathcal{I}_{\text{vis}}} \mathbf{A}^{\ell,h}[i, j],
\end{equation}
where $\mathcal{I}_{\text{txt}}$ denotes the set of instruction tokens during the prefill stage and the  generated tokens during the decode stage. A head is considered visual-sensitive if its visual attention exceeds a significance threshold:

\begin{equation}
    \mathcal{H}_{\text{vis}}^\ell = \left\{ h \,\middle|\, A_{\text{vis}}^{\ell,h}  > \mu + \lambda_{\text{vis}} \cdot \sigma \right\},
\end{equation}
where $\mu$ and $\sigma$ are the mean and standard deviation of $A_{\text{vis}}^{\ell,h}$ across all heads in layer $\ell$, and $\lambda_{\text{vis}}$ is a tunable hyperparameter controlling sensitivity. This selection identifies heads that exhibit unusually strong attention toward visual input.

\textbf{Identifying Text and System Prompt Dominant Heads}
For each attention head $h$ in layer $\ell$, we compute the total attention directed from instruction tokens to a target token set $\mathcal{C}$, defined as:

\begin{equation}
    A^{\ell,h}_{\mathcal{C}} = 
    \begin{cases}
        \sum\limits_{i \in \mathcal{I}_{\text{txt}}} \sum\limits_{j \in \mathcal{I}_{\text{txt}}} \mathbf{A}^{\ell,h}[i, j], & \text{if } \mathcal{C} = \text{txt} \\
        \sum\limits_{i \in \mathcal{I}_{\text{txt}}} \sum\limits_{j \in \mathcal{I}_{\text{sys}}} \mathbf{A}^{\ell,h}[i, j], & \text{if } \mathcal{C} = \text{sys}
    \end{cases}
\end{equation}

where $\mathcal{C} \in \{\text{txt}, \text{sys}\}$ denotes the token type of interest (i.e., text or system prompt tokens). Attention heads with dominant focus on $\mathcal{C}$ are identified by thresholding:

\begin{equation}
    \mathcal{H}^{\ell}_{\mathcal{C}} = \left\{ h \,\middle|\, A^{\ell,h}_{\mathcal{C}} > \lambda_{\mathcal{C}} \right\}
\end{equation}

This unified formulation allows HAI to selectively suppress over-attending heads based on their attention distribution patterns.
\subsubsection{Suppressing Over-Attention in System and Text Heads}
Based on Figure~\ref{fig:attention_combined}, we analyze each attention head in the LLaVA-1.5 decoder across layers for their focus on visual, textual, and system tokens. As shown in Figure~\ref{fig:enc_sys_attn_heads_distribution}, visual attention heads are sparse, while system heads are highly redundant. Figure~\ref{fig:text_attn_head} further reveals text heads that overly attend to the previous token. These patterns indicate that certain heads consistently over-focus on language priors.


To address this issue, we introduce an attention suppression mechanism that downscales attention directed to prompt-like tokens. For each decoder layer $\ell$ and head $h$, we adjust the attention matrix $\mathbf{A}^{\ell,h}$ as follows:

\begin{equation}
\mathbf{A}^{\ell,h}_{i,j} =
\begin{cases}
(1 - \alpha_{\text{txt}}) \cdot \mathbf{A}^{\ell,h}_{i,j}, & \text{if } j \in \mathcal{I}_{\text{txt}},\ h \in \mathcal{H}_{\text{txt}} \\
(1 - \alpha_{\text{sys}}) \cdot \mathbf{A}^{\ell,h}_{i,j}, & \text{if } j \in \mathcal{I}_{\text{sys}},\ h \in \mathcal{H}_{\text{sys}}
\end{cases}
\end{equation}

To maintain a valid probability distribution, the adjusted weights are re-normalized:

\begin{equation}
\mathbf{A}^{\ell,h}_{i,j} = \frac{\mathbf{A}^{\ell,h}_{i,j}}{\sum\limits_{j} \mathbf{A}^{\ell,h}_{i,j}}.
\end{equation}

where $\alpha_{\text{txt}}$ and $\alpha_{\text{sys}}$ are suppression coefficients controlling the degree of attention reduction to prior text and system prompt tokens, respectively.

\begin{table*}[htbp]
  \centering
  \small
  \caption{Comparison of F1 scores on the POPE benchmark under three evaluation settings: Random, Popular, and Adversarial. Bold indicates the best results and higher F1-score indicate better. Results are averaged over five random runs.}
  \label{tab:pope}
  \resizebox{\linewidth}{!}{ 
  \begin{tabular}{c|ccc|ccc|ccc}
    \toprule
    \multirow{2}{*}{\textbf{Methods}} & \multicolumn{3}{c|}{\textbf{LLaVA-1.5}} & \multicolumn{3}{c|}{\textbf{MiniGPT-4}} & \multicolumn{3}{c}{\textbf{mPLUG-Owl2}} \\
          & \textit{Random} $\uparrow$ & \textit{Popular} $\uparrow$ & \textit{Adversarial} $\uparrow$
          & \textit{Random} $\uparrow$ & \textit{Popular} $\uparrow$ & \textit{Adversarial} $\uparrow$
          & \textit{Random} $\uparrow$ & \textit{Popular} $\uparrow$ & \textit{Adversarial} $\uparrow$ \\
    \midrule
    Greedy        & 81.54 & 76.53 & 73.54 & 77.56 & 67.50 & 69.11 & 83.90 & 77.30 & 74.82 \\
    Beam Search   & 82.64 & 79.34 & 78.15 & 78.54 & 70.20 & 71.62 & 87.33 & 81.42 & 78.95 \\
    OPERA [CVPR2024]         & 79.50 & 76.63 & 75.88 & 78.35 & 69.65 & 71.42 & 87.03 & 80.29 & 77.92 \\
    VCD [CVPR2024]           & 82.51 & 79.33 & 78.17 & 78.61 & 69.95 & 71.62 & 87.36 & 81.42 & 78.95 \\
    DoLa [ICLR2024]          & 82.81 & 79.47 & 78.36 & 80.23 & 73.00 & 73.23 & 87.90 & 81.53 & 79.18 \\
    PAI*\footnote{w/o contrastive decoding} [ECCV2024]         & 85.94 & 81.12 & 77.75 & 78.01 & 70.26 & 72.46 & 88.18 & 81.94 & 77.83 \\
    VAR [ICLR2025]         & 81.96 & 77.40 & 73.59 & - & - & - & - & - & - \\
    \midrule
     \textbf{VisFlow (ours)}
              & \textbf{89.55} & \textbf{87.09} & \textbf{84.35}
              & \textbf{80.86} & \textbf{73.61} & \textbf{73.94}
              & \textbf{88.72} & \textbf{82.19} & \textbf{80.17} \\
    \bottomrule
  \end{tabular}
  }  
\end{table*}
\begin{table*}[htbp]
  \centering
  \small
  \caption{Comparison of CHAIR (instance-level CHAIR\textsubscript{i} and sentence-level CHAIR\textsubscript{s}) and Recall scores on the MSCOCO dataset. Smaller CHAIR\textsubscript{i} and CHAIR\textsubscript{s} indicate less hallucinations. Results are averaged over five random runs.}
  \label{tab:chair}
  \resizebox{\linewidth}{!}{
  \begin{tabular}{c|ccc|ccc|ccc}
    \toprule
    \multirow{2}{*}{\textbf{Methods}} & \multicolumn{3}{c|}{\textbf{LLaVA-1.5}} & \multicolumn{3}{c|}{\textbf{MiniGPT-4}} & \multicolumn{3}{c}{\textbf{mPLUG-Owl2}} \\
          & CHAIR\textsubscript{i} $\downarrow$ & CHAIR\textsubscript{s} $\downarrow$ & Recall $\uparrow$
          & CHAIR\textsubscript{i} $\downarrow$ & CHAIR\textsubscript{s} $\downarrow$ & Recall $\uparrow$
          & CHAIR\textsubscript{i} $\downarrow$ & CHAIR\textsubscript{s} $\downarrow$ & Recall $\uparrow$ \\
    \midrule
    Greedy       & 20.0 & 6.8 & 59.1 & 25.0 & 9.2 & \textbf{58.7} & 23.0 & 9.6 & \textbf{54.4} \\
    Beam Search  & 20.0 & 6.9 & 57.0 & 24.0 & 9.2 & 56.7 & 18.0 & 6.4 & 53.0 \\
    OPERA[CVPR2024]        & 17.0 & 6.3 & 56.7 & 20.0 & 8.2 & 58.1 & 16.0 & 5.8 & 54.0 \\
    VCD[CVPR2024]          & 20.0 & 6.9 & 57.0 & 23.0 & 8.9 & 56.4 & 18.0 & 6.4 & 53.0 \\
    DoLa[ICLR2024]         & 19.0 & 6.5 & 57.0 & 19.0 & 8.1 & 56.3 & 18.0 & 6.1 & 53.0 \\
    \midrule
    \textbf{VisFlow (ours)} 
                 & \textbf{15.0} & \textbf{3.8} & \textbf{63.1}
                 & \textbf{18.0} & \textbf{7.8} & 57.3
                 & \textbf{16.0} & \textbf{4.9} & 53.0 \\
    \bottomrule
  \end{tabular}
  }
\end{table*}

\section{Experiments}

\subsection{Implementation and Experimental Setup}

we implement VisFlow with a greedy decoding strategy (beam size = 1) and conduct all experiments on 8 NVIDIA RTX 4090 GPUs, setting a maximum generation length of 64 tokens. The hyperparameters are configured as follows: $\lambda_{\text{vis}} = 1$, $\lambda_{\text{sys}} = 0.8$, and $\lambda_{\text{txt}} = 0.3$. For LLaVA, we set $\alpha_{\text{sys}} = 0.6$ and $\alpha_{\text{txt}} = 1$. For MiniGPT4 and mPLUG-Owl2, which have shorter visual token sequences, we adjust $\alpha_{\text{sys}}$ and $\alpha_{\text{txt}}$ to 0.4 and 0.6, respectively. For the POPE dataset, we use $k = 20$ and $\delta = 20$. In contrast, for the CHAIR dataset, which necessitates fine-grained visual perception, we set $k = 10$ and $\delta = 0.4$. All baseline models are evaluated using their default settings.

To improve efficiency, attention head types are identified once at the \textit{prefill} stage and reused during decoding. TAI is applied from layer 2 onward, as early layers focus on global visual integration. HAI is applied across all layers for system heads, and to layers 0--7 for text heads to retain visual alignment and avoid over-reliance on language priors. For models with semantically compressed visual tokens (e.g., MiniGPT-4~\cite{chen2023minigpt}, mPLUG-Owl2~\cite{ye2024mplug}), TAI is omitted due to limited effectiveness.

We conducted comprehensive evaluations of VisFlow on three widely used benchmarks: CHAIR~\cite{rohrbach2018object}, POPE~\cite{li2023evaluating}, and GPT-4.1-aided evaluation~\cite{chiang2023vicuna}. These benchmarks collectively assess visual factuality, grounding robustness. Experimental results demonstrate that VisFlow consistently improves LVLM performance across diverse tasks, highlighting its effectiveness and generalizability. We compare VisFlow against several representative decoding-based methods, including VCD~\cite{leng2024mitigating}, DoLa~\cite{dola}, OPERA~\cite{huang2024opera}, PAI~\cite{liu2024paying}, and VAR~\cite{kang2025see}. Details of the benchmarks are provided in Appendix B.2.

\begin{figure}
    \centering \includegraphics[width=1\linewidth]{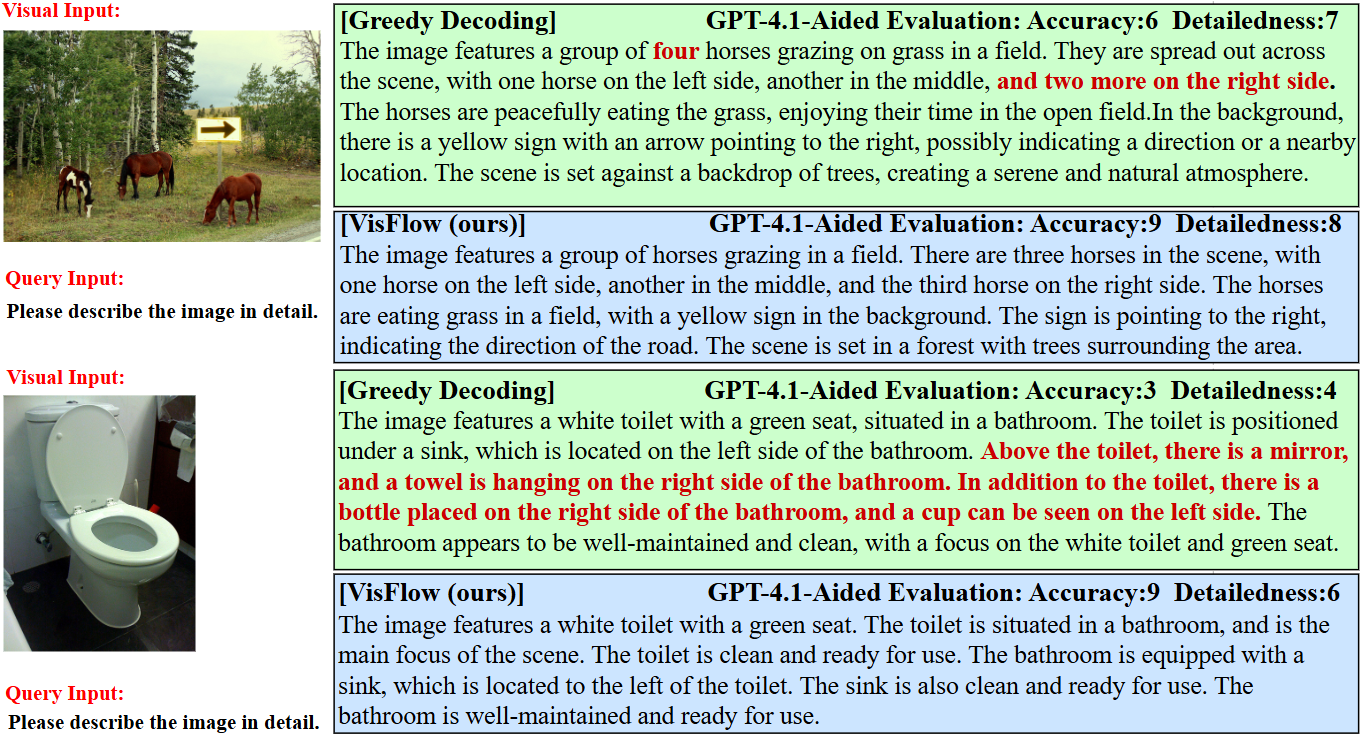}
\caption{Comparison on the MSCOCO dataset using GPT-4.1-aided evaluation. We compare greedy decoding and our proposed method (based on greedy decoding) with LLaVA-1.5. GPT-4.1 evaluations are shown alongside the responses, with hallucinated content highlighted in red.}
    \label{fig:case}
\end{figure}
\subsection{Main Results}


For the POPE benchmark, as shown in Table~\ref{tab:pope}, we achieve the highest F1 scores across all subsets. The improvements are especially notable under the Adversarial setting, highlighting VisFlow's robustness against spurious correlations. As shown in Table~\ref{tab:chair}, on the CHAIR benchmark, VisFlow yields substantially lower CHAIR\textsubscript{s} and CHAIR\textsubscript{i} scores compared to decoding-based baselines, indicating stronger alignment between generated captions and visual content. In addition, qualitative comparisons in Figure~\ref{fig:case} and GPT-4.1-aided evaluation further verify the effectiveness of our method. The human-aligned quantitative metrics provided by GPT-4.1 evaluation also corroborate VisFlow's superiority in producing faithful and grounded descriptions. Notably, the qualitative results show that our TAI effectively corrects object-count hallucinations, while our HAI mitigates language-prior-induced inertia hallucinations. More details and results on CHAIR under $L_{\text{max}} = 128$, as well as additional qualitative comparisons and GPT-4.1 evaluations, are provided in the Appendix C.





\subsection{More Analysis and Ablation Experiments}
\subsubsection{Component Ablation Study}
To assess the contribution of each component in VisFlow, we conduct ablation experiments on the CHAIR benchmark. As shown in Table~\ref{tab:ablation_visflow_components}, removing either TAI or HAI results in a noticeable drop in CHAIR scores. While removing HAI for system heads yields the lowest CHAIR\textsubscript{s}, it reduces Recall, indicating a trade-off. In contrast, our full VisFlow achieves the best overall performance, with the lowest CHAIR\textsubscript{i} and highest Recall, confirming the effectiveness of the complete design.

\begin{table}[h]
\centering
\caption{Ablation study on the CHAIR benchmark evaluating different components of VisFlow.}
\label{tab:ablation_visflow_components}
\resizebox{0.9\linewidth}{!}{%
\begin{tabular}{lccc}
\toprule
\textbf{Settings} & \textbf{CHAIR\textsubscript{s}} $\downarrow$ & \textbf{CHAIR\textsubscript{i}} $\downarrow$ & \textbf{Recall} $\uparrow$ \\
\midrule
Greedy (Baseline)               & 20.0 & 6.8 & 59.1 \\
w/o TAI                         & 16.0 & 4.8 & 56.7 \\
w/o HAI                         & 16.0 & 5.3 & 58.4 \\
w/o HAI for Txt. Heads          & 18.0 & 6.4 & 61.1 \\
w/o HAI for Sys. Heads          & \textbf{12.0} & 4.0 & 60.1 \\
\textbf{Our Full VisFlow}       & 15.0 & \textbf{3.8} & \textbf{63.1} \\
\bottomrule
\end{tabular}
}
\end{table}

\subsubsection{Sensitivity to Hyperparameters}
\begin{figure}[htbp]
    \centering
    \begin{subfigure}[b]{0.495\linewidth}
        \centering
        \includegraphics[width=\linewidth]{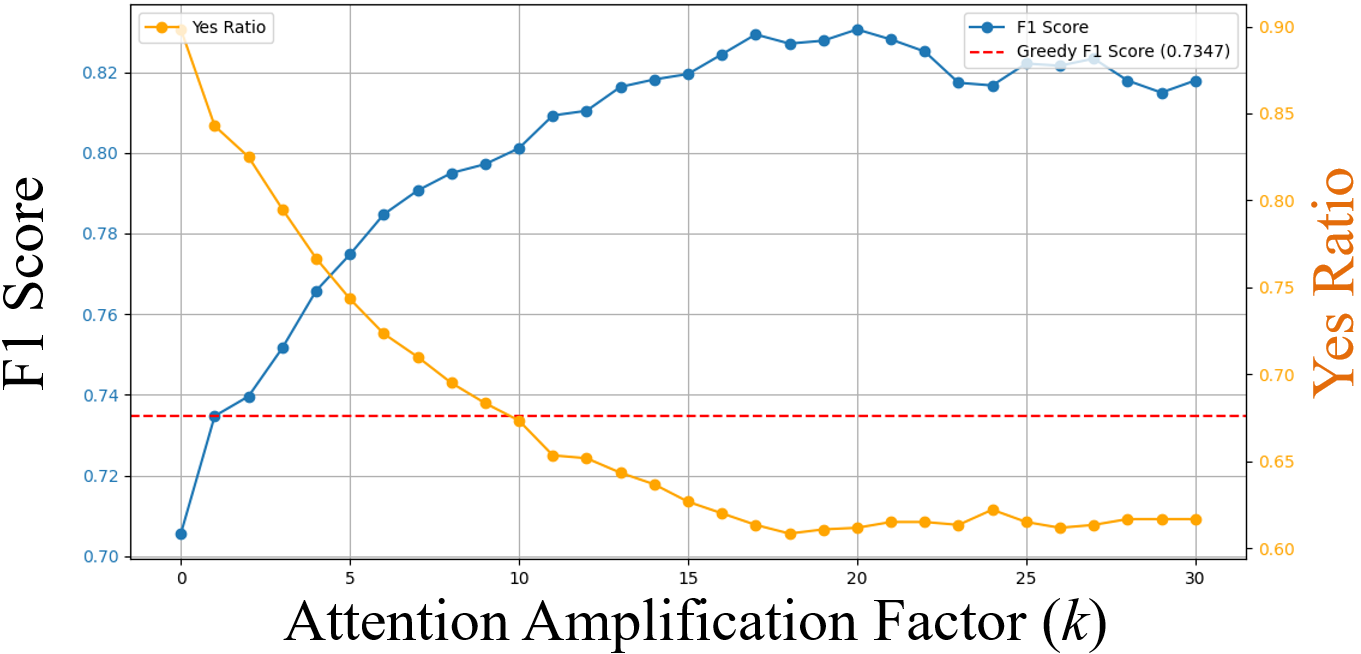}
        \caption{Effect of different enhancement factors ($k$) for salient tokens.}
        \label{fig:ablation_reallocation}
    \end{subfigure}
    \hfill
    \begin{subfigure}[b]{0.495\linewidth}
        \centering
        \includegraphics[width=\linewidth]{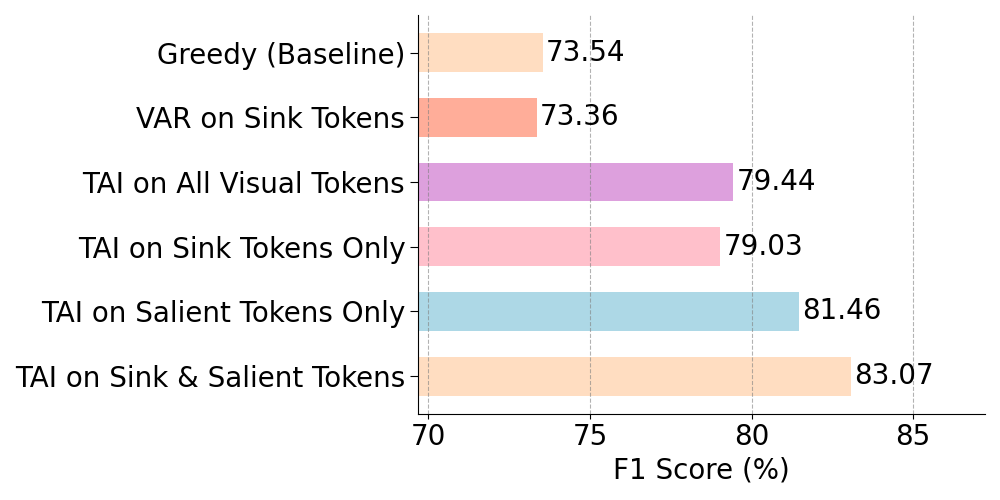}
        \caption{Comparison of token selection strategies: VAR~\cite{kang2025see} vs. TAI ($k=20$).}
        \label{fig:ablation_enhance_token}
    \end{subfigure}
    \caption{
    Ablation study of TAI on POPE (adversarial). (a) Evaluates performance sensitivity to the enhancement factor $k$; (b) compares different salient token selection methods.
    }
    \label{fig:combined_pope_ablation}
\end{figure}

To determine the optimal enhancement strength, we vary the scaling factor $k$ for salient visual tokens. As shown in Figure~\ref{fig:ablation_reallocation}, F1 Score peaks at $k=20$, while both smaller ($k=1$) and larger ($k=30$) values degrade performance, highlighting the need for balanced visual emphasis. Figure~\ref{fig:ablation_enhance_token} further shows that our selective enhancement strategy outperforms VAR~\cite{kang2025see} on visual sink tokens, supporting the claim that \textit{not all tokens are equally important}.

\begin{table}[htbp]
\centering
\caption{Comparison of masking different types of heads on the CHAIR benchmark, revealing the relative impact of visual, system, and text heads on hallucination metrics.}
\label{tab:ablation_mask_head_types}
\resizebox{0.9\linewidth}{!}{%
\begin{tabular}{lccc}
\toprule
\textbf{Setting}         & \textbf{CHAIR\textsubscript{s}} $\downarrow$ & \textbf{CHAIR\textsubscript{i}} $\downarrow$ & \textbf{Recall} $\uparrow$ \\
\midrule
Greedy (Baseline)        & 20.0                                     & 6.8                                     & \textbf{59.1} \\
Mask Visual Heads        & 46.0                                     & 22.1                                    & 56.4          \\
Mask System Heads         & 20.0                                     & 6.7                             & 58.0          \\
Mask Text Heads          & 19.0                             & 7.2                                     & 58.4          \\
Mask Text Heads (Shallow layer)          & \textbf{15.0}                             & \textbf{5.7}                                     & 57.4          \\
Mask Random Heads        & 20.0                                     & 7.4                                     & 56.0          \\
\bottomrule
\end{tabular}%
}
\end{table}
\subsubsection{Functional Analysis of Attention Heads}
To validate the functional roles of different attention head types, we identified the top-4 heads per type. We then performed causal interventions by zeroing them out from layer 0 and evaluated model performance on the CHAIR benchmark. As shown in Table~\ref{tab:ablation_mask_head_types}, masking visual-sensitive heads notably increases CHAIR scores, confirming their critical role in visual alignment. Masking text-dominant heads in shallow layer mitigates hallucinations, whereas masking system-dominant heads shows minimal impact. These findings support the claim that \textit{not all heads are equally important}.
\begin{figure}[htbp]
    \centering
    \begin{subfigure}{0.48\linewidth}
        \centering
        \includegraphics[width=\linewidth]{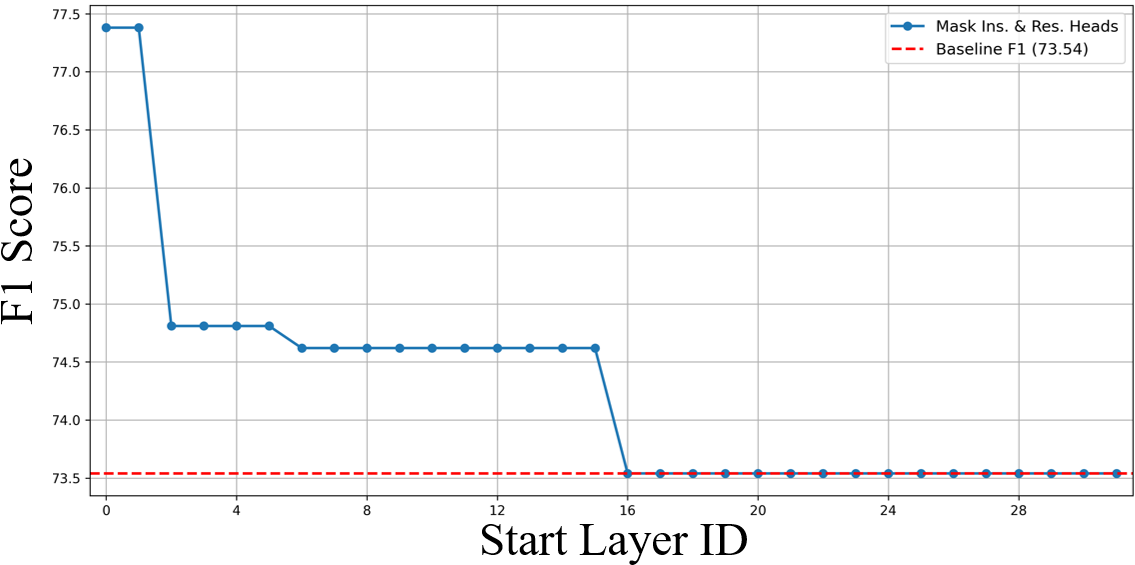}
        \caption{Masking text heads.}
        \label{fig:ablation_mask_txt_head}
    \end{subfigure}
    \hfill
    \begin{subfigure}{0.48\linewidth}
        \centering
        \includegraphics[width=\linewidth]{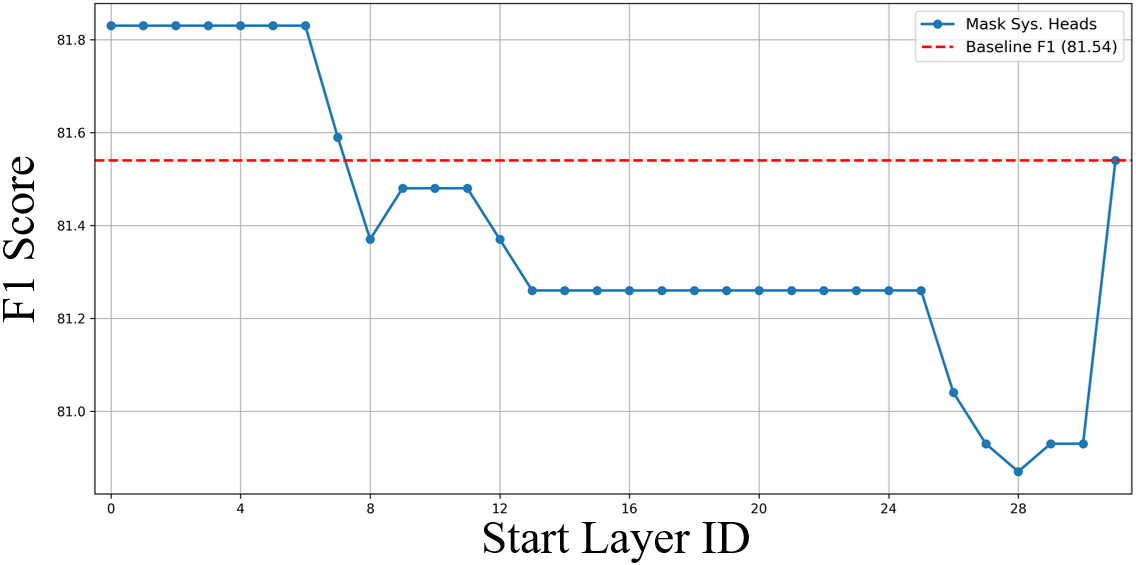}
        \caption{Masking system heads.}
        \label{fig:ablation_mask_sys_head}
    \end{subfigure}
    \caption{
    Ablation study of HAI on POPE. (a) Evaluates the effect of masking text heads on POPE (adversarial); (b) The impact of masking system heads on POPE (random).
    }
    \label{fig:ablation_mask_heads_combined}
\end{figure}
Table~\ref{tab:ablation_mask_head_types} and Figures~\ref{fig:ablation_mask_txt_head}–\ref{fig:ablation_mask_sys_head} further show POPE-based interventions with thresholds $\lambda_{\text{sys}}{=}0.8$ and $\lambda_{\text{txt}}{=}0.3$. Masking text-dominant heads effectively reduces hallucinations, while masking system heads has negligible or even positive effects, suggesting redundancy.





\subsubsection{Decoding efficiency analysis}
\begin{figure}
    \centering
\includegraphics[width=0.85\linewidth]{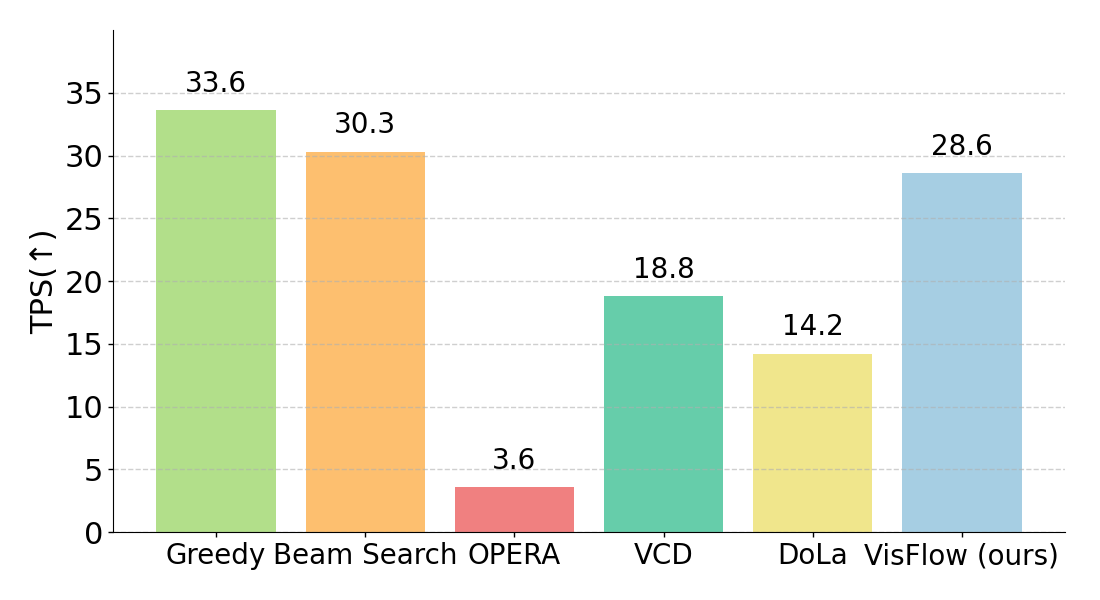}
    \caption{Comparison of our method with different baselines in terms of Tokens Per Second (TPS) during decoding on CHAIR Benchmark using LLaVA-1.5.}
    \label{fig:tps}
\end{figure}

We evaluate the inference efficiency of our method by comparing it with several baseline approaches in terms of Tokens Per Second (TPS) during decoding on the CHAIR Benchmark, as shown in Figure~\ref{fig:tps}. Our approach achieves latency comparable to beam search, while being significantly faster than other methods, demonstrating superior efficiency. 

\section{Conclusion}
In this paper, we present VisFlow, a training-free and inference-time framework for VH in LVLMs. Our framework builds on the critical insight that  \textit{not all tokens and heads are equally important for hallucination mitigation}. Based on this, VisFlow directly intervenes in the attention dynamics within the decoder to improve visual alignment and reduce reliance on linguistic priors. Specifically, we propose a dual-level attention intervention approach:
(1) TAI: Enhances attention to critical visual tokens and corrects positional biases introduced by RoPE.
(2) HAI: Suppresses over-attention to non-visual tokens. Extensive experiments on three widely-used benchmarks demonstrate that VisFlow significantly improves the visual faithfulness of generated responses. Our work provides new insights into decoding-time attention modulation as an effective means of reducing hallucination in LVLMs.


\bibliography{aaai25}

\appendix
\begin{strip}        
\section*{Technical Appendix for Not All Tokens and Heads Are Equally Important: Dual-Level Attention Intervention for Hallucination Mitigation}
\end{strip}
\section{Overview}
In this supplementary material, we present the following.
\begin{itemize}
  \item More Explanation on Motivation can be found in Section A.
  \item More Experimental Details are described in Section B.
  \item More Qualitative Results are illustrated in Section C.
\end{itemize}

\section{A. More Explanation on Motivation}
\label{sec:motivation}

To validate the motivation behind our method and assess its generalization across different LVLMs, we analyze the sparsity and layer-wise decay of visual attention from both the token- and head-level perspectives. These observations support the effectiveness of Head-Level Attention Intervention (HAI) and highlight its robustness across model variants.


\begin{figure}[H]
    \centering
    \includegraphics[width=0.8\linewidth]{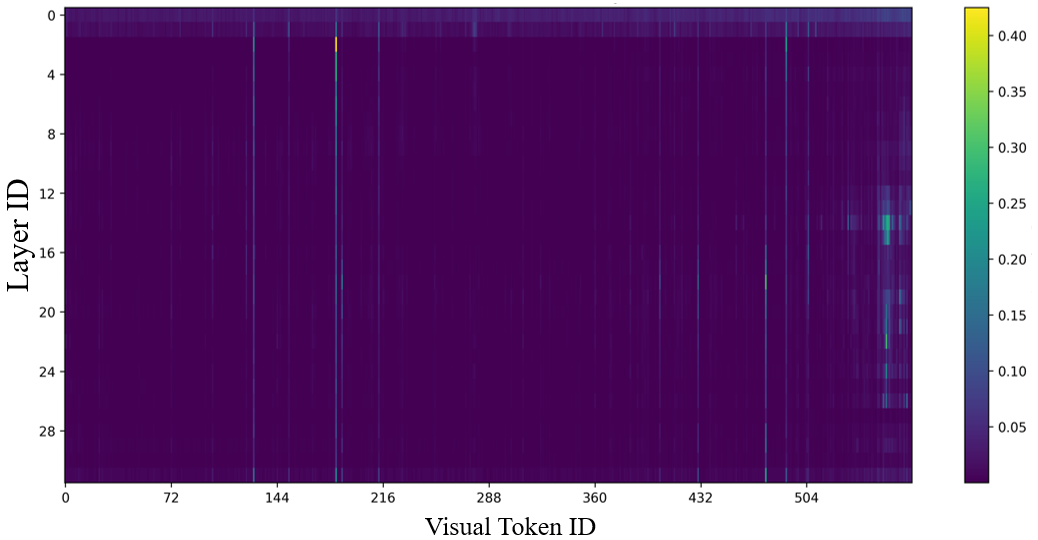}
    \caption{Layer-wise attention weights to visual tokens in LLaVA-1.5~\cite{liu2024improved}.}
    \label{fig:llava}
\end{figure}

\begin{figure}[H]
    \centering
    \includegraphics[width=0.8\linewidth]{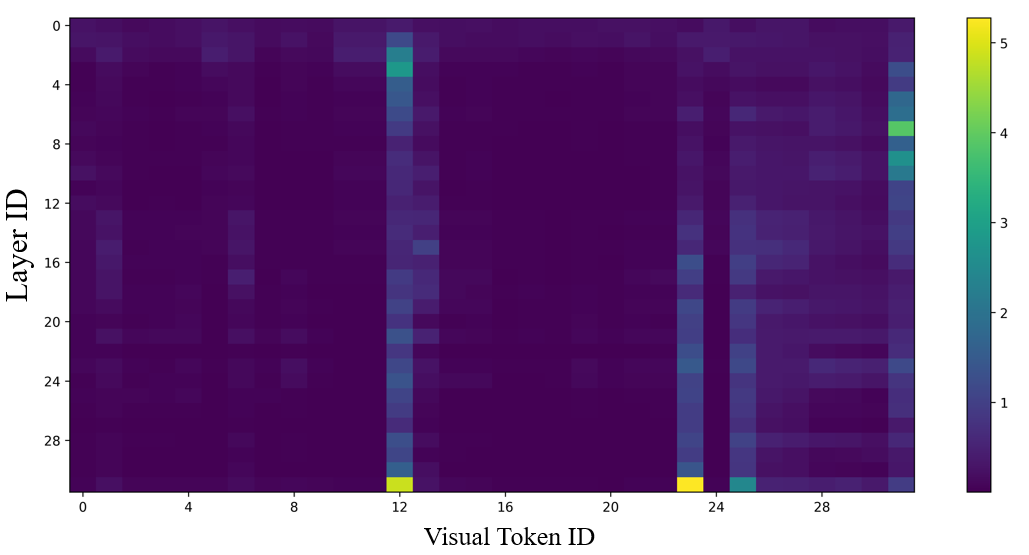}
    \caption{Layer-wise attention weights to visual tokens in MiniGPT-4~\cite{chen2023minigpt}.}
    \label{fig:minigpt}
\end{figure}

\begin{figure}[H]
    \centering
    \includegraphics[width=0.8\linewidth]{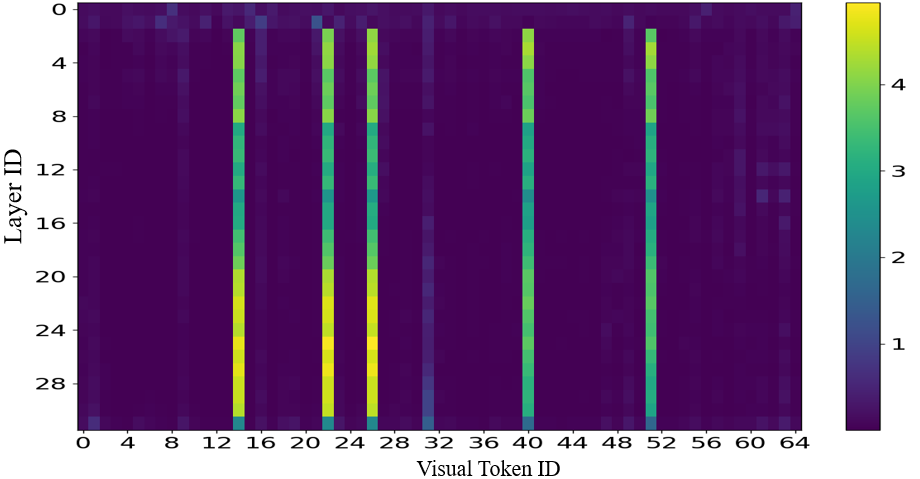}
    \caption{Layer-wise attention weights to visual tokens in mPLUG-Owl2~\cite{ye2024mplug}.}
    \label{fig:mplug}
\end{figure}

\begin{figure*}[t]
    \centering
    \begin{subfigure}{0.325\linewidth}
        \centering
        \includegraphics[width=\linewidth]{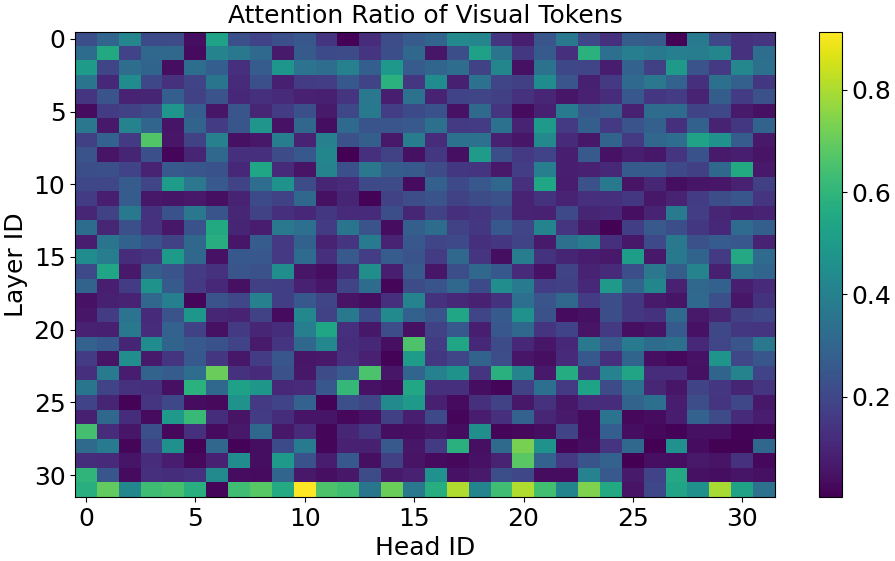}
        \caption{Visual attention}
        \label{fig:vis}
    \end{subfigure}
    \begin{subfigure}{0.30\linewidth}
        \centering
        \includegraphics[width=\linewidth]{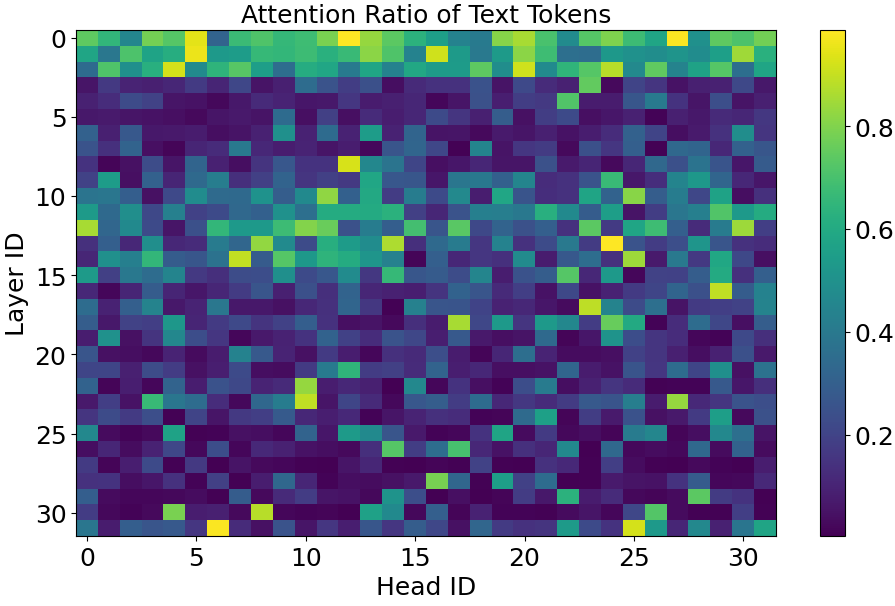}
        \caption{Text attention}
        \label{fig:text}
    \end{subfigure}
    \begin{subfigure}{0.325\linewidth}
        \centering
        \includegraphics[width=\linewidth]{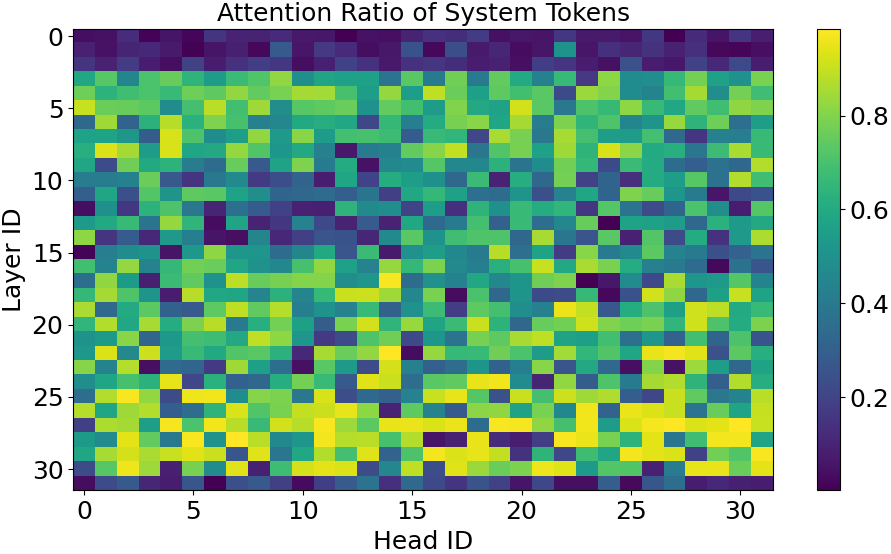}
        \caption{System attention}
        \label{fig:sys}
    \end{subfigure}
    \caption{Layer-wise heatmaps of attention weights across all heads in MiniGPT-4}
    \label{fig:minigpt_all_attn}
\end{figure*}
\begin{table*}[t]
\centering
\renewcommand{\arraystretch}{1.2}
\begin{tabular}{lccc}
\toprule
\textbf{Model} & \textbf{System Prompt Indices} & \textbf{Visual Token Indices} & \textbf{Text Token Indices} \\
\midrule 
\textbf{LLaVA-1.5}~\citep{liu2024improved}   & [0:35]   & [35:611]   & [611:] \\
\textbf{MiniGPT-4}~\citep{chen2023minigpt}   & [0:7]    & [7:39]     & [39:]  \\
\textbf{mPLUG-Owl2}~\citep{ye2024mplug}      & [1:5]    & [5:70]     & [70:]  \\
\bottomrule
\end{tabular}
\caption{Token index ranges for four token types across LVLMs. Instruction and response tokens are listed separately.}

\label{tab:token_indices}
\end{table*}
As illustrated in Figures~\ref{fig:llava}--\ref{fig:mplug}, a consistent trend is observed across models: 
\vspace{-0.5em}
\begin{itemize}
    \item \textbf{Sparse Visual Attention:} The attention weights assigned to visual tokens are generally sparse, indicating that only a few visual tokens receive significant attention. Moreover, the attention heads that actively attend to visual information are also sparse, as demonstrated in~\ref{fig:vis}.
    \item \textbf{Layer-wise Decay:} As the decoding layer deepens, the overall attention to visual tokens diminishes significantly. This suggests that visual information becomes less involved in later stages of generation.
    \item \textbf{RoPE-induced Positional Bias:} We observe a positional skew, where the visual tokens closer to the instruction text receive disproportionately higher attention. This is due to the Rotary Position Embedding (RoPE)~\cite{su2024roformer}, which introduces a positional bias based on relative distance in the sequence. As a result, visual tokens appearing later in the sequence (i.e., those with positional indices, closer to the text) dominate attention.
\end{itemize}
\vspace{-0.5em}

These findings highlight a critical limitation in current LVLMs: visual inputs are neither uniformly nor sufficiently integrated during text generation, which compromises the model’s ability to fully leverage visual cues. To address this, our method introduces Token-level Attention Intervention (TAI) to reinforce attention to salient visual tokens and correct positional biases introduced by RoPE. In parallel, Head-level Attention Intervention (HAI) mitigates the dominance of language priors and further enhances the model’s visual grounding capability.
\section{B. More Experimental Details }
\label{sec:exp_detail}

\subsection{B.1. Division of Multimodal Tokens}
We divide the input into four token types: system prompts, visual tokens, instruction tokens, and response tokens. Their index ranges in different LVLMs are listed in Table~\ref{tab:token_indices}. Due to the use of Q-former~\cite{li2023blip}, MiniGPT-4~\cite{chen2023minigpt} and mPLUG-Owl2~\cite{ye2024mplug} have significantly fewer visual tokens than LLaVA-1.5~\cite{liu2024improved}.

\subsection{B.2. Benchmarks and Metrics}

We conduct extensive experiments on the following benchmarks:

\begin{itemize}
    \item \textbf{POPE~\citep{li2023evaluating}} is designed to measure object hallucination in LVLMs. Each image is paired with six yes/no questions regarding the presence of specific objects, based on images from MSCOCO dataset~\cite{lin2014microsoft}. Negative samples are constructed in \textit{random}, \textit{popular}, and \textit{adversarial} fashions. The instruction text provided to the models is: \textit{``Find evidence first and then answer: is there a \{object\} in the image?''} In total, POPE includes 3,000 test cases, and performance is evaluated using accuracy, precision, recall, and F1 score.

    \item \textbf{CHAIR~\citep{rohrbach2018object}} focuses on object hallucinations in image captioning tasks. LVLMs are instructed with: \textit{``Please describe this image in detail.''} The evaluation uses 500 randomly selected MSCOCO validation images. Two main metrics are reported: $\text{CHAIR}_i$ and $\text{CHAIR}_s$, calculated as
    \begin{align}
        \text{CHAIR}_i &= \frac{|\text{hallucinated objects}|}{|\text{all objects mentioned}|}, \\
        \text{CHAIR}_s &= \frac{|\text{captions with hallucinated object}|}{|\text{all sentences}|},
    \end{align}
    where $\text{CHAIR}_i$ measures the proportion of hallucinated objects among all mentioned objects, and $\text{CHAIR}_s$ captures the fraction of sentences containing at least one hallucinated object.
    \item \textbf{GPT-4.1 Assisted Benchmarks\footnote{\url{https://huggingface.co/datasets/liuhaotian/llava-bench-in-the-wild}}} are used to evaluate models' fine-grained visual hallucination (VH) performance. This benchmark contains 60 carefully designed questions targeting nuanced understanding and hallucination detection. Model responses are assessed by GPT-4.1~\cite{chiang2023vicuna}, which serves as an advanced vision-language evaluator, measuring both factual accuracy and descriptive detail. The prompt format used to elicit consistent judgments from GPT-4.1 is provided in Table~\ref{tab:gpt4_eval_prompt}.
\end{itemize}

\begin{table*}[htbp]
\centering
\renewcommand{\arraystretch}{1.3}
\begin{tabularx}{\textwidth}{>{\raggedright\arraybackslash}X}
\toprule
\textbf{Description:} \\
AI that scores image description accuracy and detailedness. \\
\midrule
\textbf{Instructions:} \\
You are an AI designed to evaluate and score the performance of two AI assistants in describing a given image. Your primary focus is on the accuracy and detailedness of their descriptions. You will assess the accuracy by checking for hallucinations—any part of the description that is inconsistent with the image content. For detailedness, you will consider how rich the response is in necessary details, excluding any hallucinated parts. You will provide scores on a scale from 1 to 10 for each assistant separately, based on these criteria. After scoring, you will offer an explanation for your evaluation, ensuring it is free from bias and not influenced by the order of presentation of the responses. \\
\midrule
\textbf{Input format:} \\
\texttt{[Assistant 1]} \\
\texttt{\{Response 1\}} \\
\texttt{[End of Assistant 1]} \\
\texttt{[Assistant 2]} \\
\texttt{\{Response 2\}} \\
\texttt{[End of Assistant 2]} \\
\midrule
\textbf{Output format:} \\
\texttt{Accuracy:} \\
\texttt{Scores of the two answers:} \\
\texttt{Reason:} \\
\texttt{Detailedness:} \\
\texttt{Scores of the two answers:} \\
\texttt{Reason:} \\
\bottomrule
\end{tabularx}
\caption{GPT-4.1-aided evaluation setup. This prompt is used to guide GPT-4.1 in evaluating LVLM-generated responses based on accuracy (absence of hallucinations) and detailedness.}
\label{tab:gpt4_eval_prompt}
\end{table*}

\section{C. More Qualitative Results}
\label{sec:results}

\subsection*{C.1. Visualization Token-Level Attention Intervention}

To comprehensively validate the effectiveness of our Token-Level Attention Intervention (TAI), we provide both a direct comparison of visual attention distributions before and after TAI, as well as detailed visualizations of identified salient tokens.

\begin{figure}[h]
    \centering
    \includegraphics[width=1\linewidth]{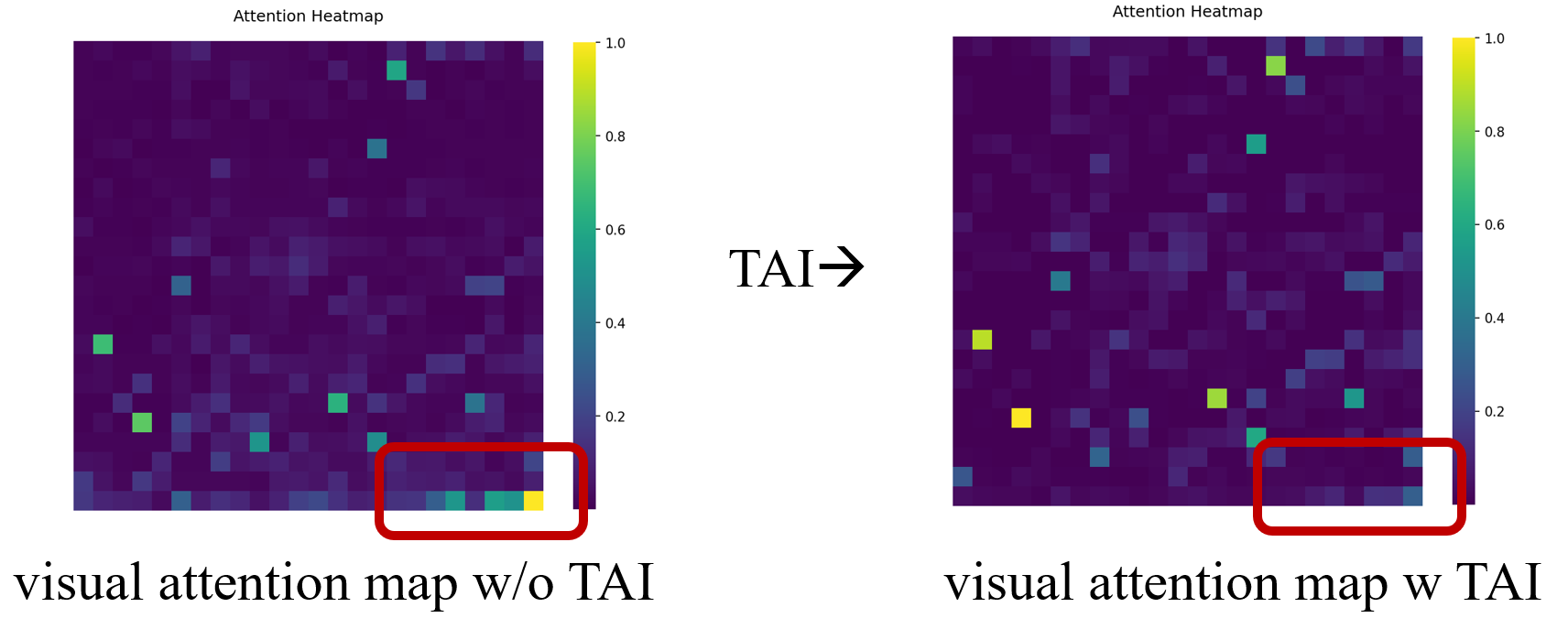}
    \caption{Comparison of visual attention distributions before and after TAI intervention. The figure demonstrates that our proposed TAI effectively alleviates the visual attention bias caused by RoPE, leading to more accurate and semantically meaningful attention allocation.}
    \label{fig:de_rope}
\end{figure}

Figure~\ref{fig:de_rope} visualizes the distribution of visual attention before and after applying TAI. This comparison highlights that TAI effectively alleviates the visual attention bias caused by RoPE, leading to more accurate and semantically meaningful attention allocation. The TAI intervention redistributes attention from biased or irrelevant regions to semantically significant areas, thus supporting improved visual alignment and model interpretability. This result further demonstrates the necessity and effectiveness of TAI in addressing the inherent shortcomings of RoPE within LVLM frameworks.

Figure~\ref{fig:three_subfigs} presents visualization examples of visual salient token identification in LLaVA-1.5 7B. For each input image, we display the tokens recognized as \textit{Visual Salient Tokens}, which correspond to meaningful visual regions essential for accurate grounding. These salient tokens differ from visual sink tokens that may attract substantial attention but lack semantic relevance.

The visualized cases demonstrate that our TAI accurately and effectively targets these critical visual regions. By focusing on the salient tokens, TAI enhances model \textit{interpretability} and supports more precise vision-language alignment. These examples confirm the specificity and efficacy of our approach in isolating regions genuinely necessary for downstream tasks.


\begin{figure}[h]
    \centering
    \includegraphics[width=0.7\linewidth]{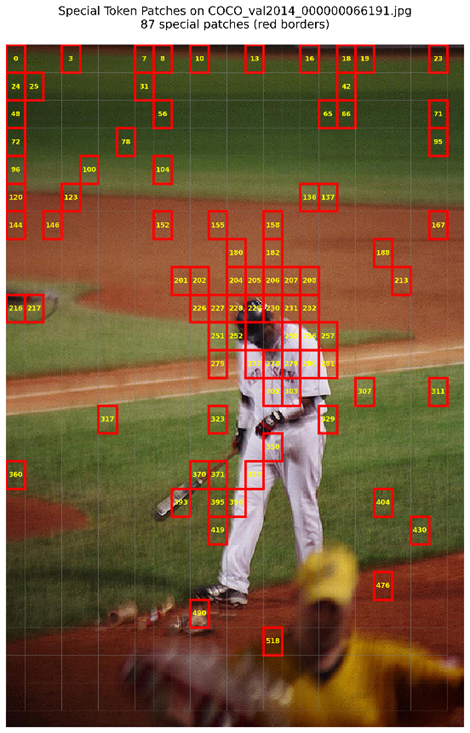}
    \includegraphics[width=0.7\linewidth]{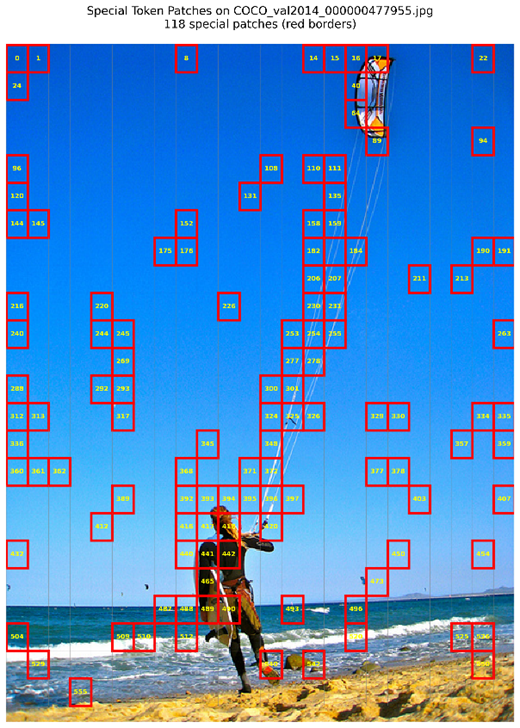}
    \includegraphics[width=0.7\linewidth]{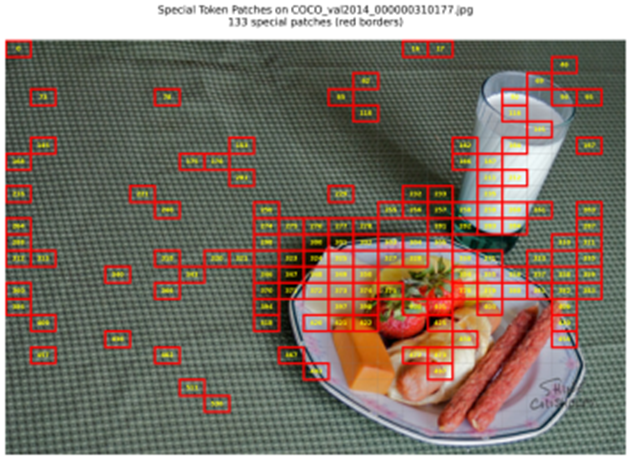}
    \caption{Visualization of visual attention in LLaVA-1.5-7B. (a), (b), and (c) showcase the identified \textbf{Visual Salient Tokens}: tokens that align with meaningful visual regions critical for grounding. These results verify the targeted and effective intervention of TAI in enhancing visual alignment.}
    \label{fig:three_subfigs}
\end{figure}

As illustrated in Figure~\ref{fig:three_subfigs}, the highlighted tokens correspond to semantically important regions of the images, validating that our TAI approach robustly identifies and reinforces the alignment of visual salient information. Additional visualization cases are provided in the Appendix for further reference.
\subsection{C.2. CHAIR Benchmark Results with $L_{\text{max}} = 128$}

More details and results on the CHAIR benchmark are provided in this section. We use the LLaVA-1.5 7B model as a representative example and evaluate it under the standard decoding strategy, greedy decoding (beam size = 1), with a maximum generation length $L_{\text{max}} = 128$ tokens. 

We compare three decoding strategies:
\begin{itemize}
    \item \textbf{Greedy Decoding}  The baseline method that selects the token with the highest probability at each step.
    \item \textbf{VCD (Visual Consistency Decoding)}~\cite{leng2024mitigating} reduces hallucination by contrasting against distributions prone to generating more hallucinations, thereby enabling better correction during generation.
    \item \textbf{VisFlow (Ours)}  Our proposed decoding strategy, which further improves the factual consistency and alignment between generated text and visual content.
\end{itemize}

\begin{figure*}
    \centering
    \includegraphics[width=0.95\linewidth]{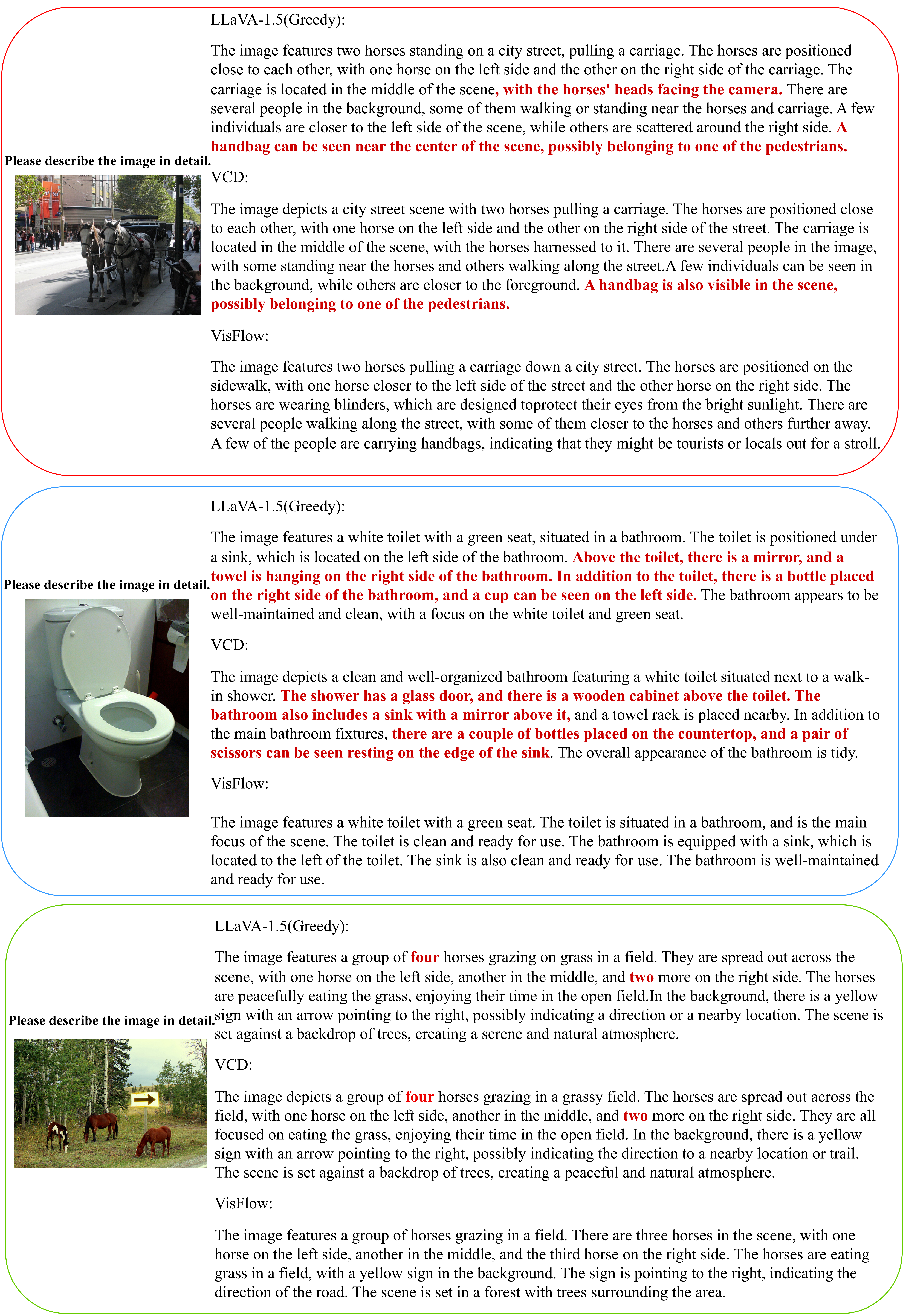}
    \caption{More cases on the CHAIR benchmark with $L_{\text{max}} = 128$. Hallucinated content is highlighted in red.}
    \label{fig:case}
\end{figure*}

Qualitative examples are illustrated in Figure~\ref{fig:case}, with hallucinated content highlighted in red. These results demonstrate the effectiveness and superiority of our method in reducing hallucinations and improving grounding accuracy under the CHAIR benchmark.

\end{document}